\documentclass[10pt,journal,compsoc]{IEEEtran}

\ifCLASSOPTIONcompsoc
  \usepackage[nocompress]{cite}
\else
  \usepackage{cite}
\fi
\usepackage[T1]{fontenc}
\usepackage{graphicx}
\usepackage{amsmath}
\usepackage{url}
\usepackage{amssymb}
\usepackage{booktabs}
\usepackage{tikz}
\usepackage{etoolbox}
\usepackage{ragged2e}
\usepackage{enumitem}
\usepackage{caption}
\usepackage{multirow}
\newcommand*{\circled}[1]{\lower.7ex\hbox{\tikz\draw (0pt, 0pt)%
    circle (.5em) node {\makebox[1em][c]{\small #1}};}}
\robustify{\circled}
\usepackage[colorlinks,linkcolor=blue]{hyperref}

\newcommand{\ba}{\mathbf{a}}
\newcommand{\bc}{\mathbf{c}}

\newcommand{\bd}{\mathbf{d}}

\newcommand{\bsd}{\mathbf{s'}}

\newcommand{\hbc}{\hat{\bc}}
\newcommand{\hba}{\hat{\ba}}
\newcommand{\hbd}{\hat{\bd}}

\newcommand{\hbsd}{\hat{\bsd}}

\newcommand{\OURS}{3DETR\xspace}

\renewcommand{\paragraph}[1]{\vspace{-1.5mm}{\flushleft\textbf{#1}}}

\newcommand{\updateblue}[1]{\textcolor{black}{#1}}
\usepackage{caption}

\graphicspath{{images_paper/}}
\newcommand{\figref}[1]{Fig.~\ref{#1}}
\newcommand{\tabref}[1]{Tab.~\ref{#1}}
\newcommand{\secref}[1]{Sec.~\ref{#1}}

\newcommand{\equref}[1]{Eqn.~(\ref{#1})}
\usepackage{booktabs}
\usepackage{overpic}
\usepackage{xcolor}
\usepackage{subfig}
\usepackage{amsfonts}
\usepackage{amsmath}
\usepackage{graphicx}
\usepackage{float}
\usepackage{booktabs}
\usepackage{wrapfig}
\usepackage[ruled,vlined]{algorithm2e}

\ifCLASSINFOpdf
\else
\fi

\begin{document}

\title{Collaborative Novel Object Discovery and Box-Guided Cross-Modal Alignment for Open-Vocabulary 3D Object Detection}

\author{Yang Cao,~Yihan Zeng,~Hang Xu,~and~Dan~Xu~\IEEEmembership{Member,~IEEE}
\IEEEcompsocitemizethanks{\IEEEcompsocthanksitem Yang Cao and Dan Xu are with the Department
of Computer Science and Engineering, The Hong Kong University of Science and Technology, Hong Kong SAR.
\protect
E-mail: \{ycaobd, danxu\}@cse.ust.hk
\IEEEcompsocthanksitem Dan Xu is the corresponding author.
\IEEEcompsocthanksitem Yihan Zeng and Hang Xu are with the Huawei Noah's Ark Lab.
}%
}

\markboth{Journal of \LaTeX\ Class Files,~Vol.~14, No.~8, August~2015}%
{Shell \MakeLowercase{\textit{et al.}}: Bare Demo of IEEEtran.cls for Computer Society Journals}

\IEEEtitleabstractindextext{%
\begin{abstract}
\justifying
Open-vocabulary 3D Object Detection (OV-3DDet) addresses the detection of objects from an arbitrary list of novel categories in 3D scenes, which remains a very challenging problem. In this work, we propose \textbf{CoDAv2}, a unified framework designed to innovatively tackle both the localization and classification of novel 3D objects, under the condition of limited base categories. For localization, the proposed 3D Novel Object Discovery~(3D-NOD) strategy utilizes 3D geometries and 2D open-vocabulary semantic priors to discover pseudo labels for novel objects during training. 3D-NOD is further extended with an Enrichment strategy that significantly enriches the novel object distribution in
the training scenes,
and then enhances the model's ability to localize more novel objects. 
The 3D-NOD with Enrichment is termed
3D-NODE.
For classification, the Discovery-driven Cross-modal Alignment (DCMA) module aligns features from 3D point clouds and 2D/textual modalities, employing both class-agnostic and class-specific alignments that are iteratively refined to handle the expanding vocabulary of objects. Besides, 2D box guidance boosts the classification accuracy against complex background noises, which is coined as Box-DCMA. Extensive evaluation demonstrates the superiority of CoDAv2. CoDAv2 outperforms the best-performing method by a large margin~($\text{AP}_{Novel}$ of \textbf{9.17 vs. 3.61} on SUN-RGBD and \textbf{9.12 vs. 3.74} on ScanNetv2). Source code and pre-trained models are available at the~\href{https://github.com/yangcaoai/CoDA_NeurIPS2023}{GitHub project page}.

\end{abstract}

\begin{IEEEkeywords}
Open-vocabulary 3D object detection;
Multi-modality learning; 3D perception;
\end{IEEEkeywords}}

\maketitle

\IEEEdisplaynontitleabstractindextext

\IEEEpeerreviewmaketitle

\IEEEraisesectionheading{\section{Introduction}\label{sec:introduction}}
\IEEEPARstart{T}{he} field of computer vision extensively explores the task of 3D object detection (3D-Det)~\cite{misra2021end,qi2019deep,qi2017pointnet}, which is crucial for applications such as industrial automation, robotic systems, and autonomous driving. This task involves localization and classification of objects within 3D point cloud environments. 
Nevertheless, the potential application of most existing methods is restricted as they heavily depend on datasets comprised of fixed and known categories. This limitation hampers their utility in diverse real-world scenarios, where one encounters more novel and variable object categories.
The emerging field of {Open-vocabulary 3D object detection (OV-3DDet)}~\cite{lu2023open,cao2023coda} aims to address this issue by enabling the detection of novel object types. In OV-3DDet, the approach involves training models on datasets with limited base categories, yet these models are evaluated in environments populated with novel objects. This discrepancy poses significant challenges, particularly in how to localize and classify these novel objects with models trained only on the annotations of very limited known base categories.
To date, only a few studies have tackled this demanding aspect of OV-3DDet, despite its importance in broadening the applicability of 3D object detection systems.
\par To capture novel object boxes, OV-3DET~\cite{lu2023open} harnesses a well-established pre-trained 2D open-vocabulary object detection model (OV-2DDet)~\cite{zhou2022detecting}. This model is used to identify a wide array of 2D novel object boxes, which facilitates the generation of pseudo 3D object box labels for the respective novel 3D objects. 
Instead of depending on an external OV-2DDet model to provide many annotations of
novel objects for training the model,
our method~\cite{cao2023coda} seeks to learn a discovery of novel 3D object boxes, based on limited 3D annotations of base categories on the training data.
Furthermore, using both the human-annotated base object boxes and the newly discovered novel object boxes, our approach explores joint class-agnostic and class-specific cross-modal alignment. This integrated strategy of novel object discovery and cross-modal alignment aims to collaboratively improve both the localization and classification of novel objects simultaneously.

\begin{figure}[!t]
\centering
    \begin{overpic}[width=1\linewidth]{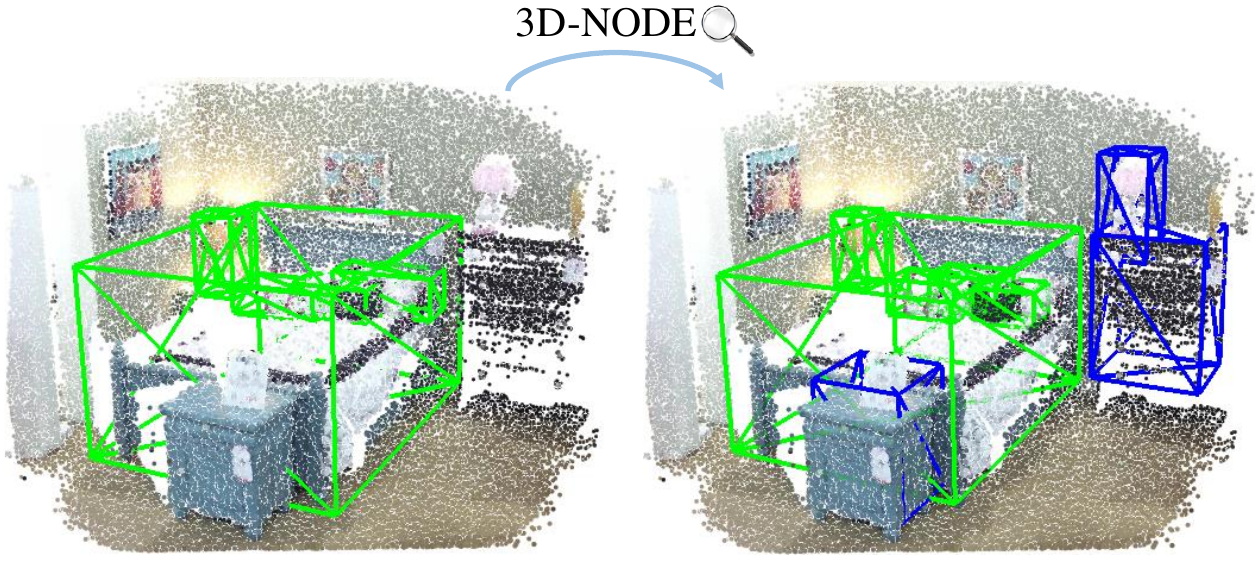}
    \end{overpic}
    \caption{Our 3D Novel Object Discovery with Enrichment~(3D-NODE) encourages the model to detect more novel objects, e.g.,
the dresser and night stand marked by blue
boxes.
}\label{fig:3D_NODE}
\end{figure}
Specifically, to enhance the localization of novel objects, we develop a training strategy termed as~\textbf{3D Novel Object Discovery} (3D-NOD). This strategy is designed to leverage information from both the 3D and 2D domains, to effectively discover novel objects throughout the training stage. Our approach trains the model to predict class-agnostic 3D bounding boxes using annotations from base category 3D boxes, thereby enabling the model to derive 3D geometry priors from the geometric characteristics of 3D point clouds.
Additionally, to confirm the discovered novel boxes, we employ the pre-trained vision-language model CLIP~\cite{radford2021learning}, which is adapted to provide 2D semantic category priors.
This dual consideration of both 3D and 2D clues allows for the effective discovery of novel objects during iterative training.
Further, we observe that the distribution of data typically follows a long-tail pattern, which means that the number of samples for novel object categories in the training set is relatively low. This scarcity poses significant challenges for the model to effectively learn to detect novel objects. To address this issue, we propose \textbf{3D novel object enrichment} for the extension of 3D-NOD.
Specially, during training, we retain the novel objects discovered by 3D-NOD in a novel object data pool. By randomly sampling and inserting these discovered novel objects into the input 3D scenes during training, we significantly enhance the presence of novel objects. 
This augmentation strategy enriches the novel object distribution in the training scenes, and then encourages the model to detect more novel objects. The proposed strategies of \textbf{3D} \textbf{N}ovel \textbf{O}bject \textbf{D}iscovery and \textbf{E}nrichment are collectively termed as \textbf{3D-NODE}, which enriches the novel objects at both the label and data levels. As illustrated in~\figref{fig:3D_NODE}, the model successfully detects more novel objects benefiting from the proposed 3D-NODE. 

\par To achieve the novel object classification, building upon 3D-NOD,
we further propose a \textbf{Discovery-Driven Cross-Modal Alignment} (DCMA) module. This module is designed to align 3D object features with 2D object and text features in a large vocabulary space. It effectively utilizes both human-annotated 3D boxes from base categories and the 3D boxes of novel categories that were discovered. The DCMA module comprises two major components: category-agnostic distillation and class-specific contrast alignment.
The category-agnostic distillation aligns 3D point cloud features with 2D CLIP image features, without depending on specific category labels. This distillation process is designed to push the features of the 3D point clouds and the 2D images closer
wherever a 3D box covers.
On the other hand, DCMA focuses on aligning 3D object features with CLIP text features within a large category vocabulary space. This alignment uses category-specific information from both the annotated base objects and the discovered novel objects, employing a cross-modal contrastive learning manner.
Furthermore, we find that CLIP~\cite{radford2021learning} performs unsatisfactorily in background regions. Given that the open-vocabulary classification ability of our method is derived from CLIP, our method may mistakenly interpret background regions as object regions. To improve discrimination capability for background regions, we have enhanced our DCMA by incorporating 2D box guidance.
In detail, we first prepare 2D guidance boxes, which are generated from an open-vocabulary 2D detector~\cite{chen2023sharegpt4v,liu2024grounding}, and cover most of the foreground regions. Subsequently, the predicted 3D boxes from our model are projected onto the 2D image plane. We then calculate the IoU between the projected 2D boxes and the 2D guidance boxes to determine if the predicted boxes cover background regions. This procedure is named \textit{2D-box-guide background matching}.
Boxes identified as matching the background are aligned with the `background' category during the subsequent cross-modal alignment phase of training. Thus, with the aid of 2D box guidance, our method more effectively discriminates between foreground objects and background areas. This box-guided cross-modal alignment is coined as \textbf{Box-DCMA}.

This paper significantly extends our previous NeurIPS version,
i.e., CoDA~\cite{cao2023coda}. Specifically, we further propose CoDAv2 by extending the two key components of CoDA: 3D-NOD and DCMA.
First, we extend 3D-NOD to 3D-NODE by proposing 3D object Enrichment.
In detail, during training, we maintain a pool of novel objects discovered by 3D-NOD. By randomly sampling from this pool and inserting the novel objects into the input scenes, we significantly increase the occurrence of novel objects. This augmentation strategy enriches the distribution of novel objects in the training scenes,
thereby achieving clearly enhanced detection of novel objects.
Besides, we extend our DCMA to Box-DCMA by integrating 2D box guidance. We begin by preparing 2D guidance boxes, which are generated by an open-vocabulary 2D detector~\cite{chen2023sharegpt4v,liu2024grounding} and cover the majority of the foreground regions, 
which serve as important guidance in the following matching process.
The predicted 3D boxes from our model are then projected onto the 2D image plane. We proceed to compute the IoU between these projected 2D boxes and the 2D guidance boxes to assess whether the predicted boxes cover background areas. 
During the subsequent cross-modal alignment phase of training, these matched background boxes are specifically aligned with the `background' category. By leveraging the 2D box guidance, our method can more effectively distinguish between foreground objects and background regions, enhancing the discriminative capability of our model. 
Benefiting from both 3D-NODE and Box-DCMA, our extended framework in this paper, named \textbf{CoDAv2}, significantly outperforms the best-performing alternative method by more than \textbf{140\%} on two challenging datasets,
i.e., SUN-RGBD and ScanNetv2.

In summary, the key contributions of this paper are threefold:
\begin{itemize}[leftmargin=*]
\item We propose an end-to-end open-vocabulary 3D object detection framework, termed as \textbf{CoDA}, designed to simultaneously localize and classify novel objects. We subsequently evolve this framework into \textbf{CoDAv2}, which boosts both localization and classification capabilities. CoDAv2 significantly surpasses the most competitive alternative method by a substantial margin, achieving $\text{AP}_{Novel}$ scores of \textbf{9.17 vs. 3.61} on SUN-RGBD and \textbf{9.12 vs. 3.74} on ScanNetv2.
\item We develop an effective 3D Novel Object Discovery (\textbf{3D-NOD}) strategy capable of localizing novel objects through joint utilization of 3D geometry and 2D semantic open-vocabulary priors. Based on objects discovered through this strategy, we further design the Discovery-Driven Cross-Modal Alignment (\textbf{DCMA}), which can align 3D point cloud features with 2D image/text features across a broad vocabulary spectrum.
\item We extend 3D-NOD with the 3D Novel Object Enrichment strategy, which involves integrating discovered novel objects into the training scenes to actively promote the detection of more novel objects. The proposed strategies of 3D Novel Object Discovery with Enrichment are collectively termed \textbf{3D-NODE}, which enriches the
novel objects at both the label and data levels. Furthering the capabilities of DCMA, we also introduce box guidance to improve the ability to distinguish between background and foreground objects. This advancement is named the discovery-driven cross-modal alignment module with box guidance (\textbf{Box-DCMA}).
\end{itemize}

\section{Related Works}\label{sec:related}
\par\noindent\textbf{3D Object Detection.} Recent advancements have propelled 3D object detection (3D-Det) \cite{qi2019deep,xie2020mlcvnet, zhang2021learning, ma2021delving, zhang2020h3dnet, xie2021venet, cheng2021back} to new heights. A notable innovation is VoteNet\cite{qi2019deep}, which integrates a point voting system in 3D-Det. This system utilizes PointNet~\cite{qi2017pointnet} for processing 3D points, grouping them into clusters from which object features are subsequently extracted for prediction purposes. Expanding upon this, MLCVNet~\cite{xie2020mlcvnet} has developed enhanced point voting modules that effectively gather and utilize multi-level contextual data to improve detection accuracy. 
The advent of transformer-based technologies~\cite{carion2020end} has further enriched the landscape of 3D-Det. For instance, GroupFree~\cite{liu2021group} employs a transformer to serve as the predictive head, bypassing the traditional need for manual grouping. Furthermore, 3DETR~\cite{misra2021end} introduces a pioneering end-to-end transformer-based framework specifically designed for 3D object detection. 
\updateblue{STEMD~\cite{zhang2024spatial} significantly enhances the DETR paradigm for multi-frame 3D object detection by introducing a spatial-temporal graph attention network to model inter-object interactions and temporal dependencies, leveraging previous frame outputs for improved query initialization, and employing an IoU regularization term to reduce redundant predictions.}
However, these methodologies typically confine themselves to a close-vocabulary context, where the categories of objects are predefined and remain unchanged throughout the training and testing phases. Diverging from this norm, our research aims at open-vocabulary 3D-Det through the development of a comprehensive system known as CoDA and CoDAv2, facilitating the simultaneous discovery and classification of novel objects.

\par\noindent\textbf{Open-Vocabulary 2D Object Detection.}
Turning to 2D perception~\cite{he2017mask,girshick2015fast,redmon2016you,carion2020end,ye2022taskprompter,zhao2019contrast,ye2023seggen,ye2022inverted,pi2023detgpt}, the field of open-vocabulary 2D object detection (OV-2DDet) has expanded significantly, as evidenced by numerous studies~\cite{feng2022promptdet,gu2021open,gupta2022ow,radford2021learning,jia2021scaling,yaodetclip,minderer2022simple,du2022learning,ma2022open,zhou2022detecting,zareian2021open,yao2023detclipv2,yao2024detclipv3} pushing its boundaries. Notably, RegionCLIP~\cite{zhong2022regionclip} elevates the CLIP model from a mere image-level application to a more granular regional level, enabling precise alignment between regional visuals and textual features. In contrast, Detic~\cite{zhou2022detecting} leverages image-level annotations from ImageNet~\cite{deng2009imagenet} to assign contextual labels to object proposals. Additionally, approaches like GLIP~\cite{li2022grounded} and MDETR~\cite{kamath2021mdetr} conceptualize detection as a text-based grounding exercise, utilizing textual queries to identify and delineate objects within images based on specified descriptors. 
While these innovations advance the capabilities of 2D detection, our project redirects focus toward the more intricate and less frequented domain of open-world 3D object detection, a challenge that has yet to be thoroughly explored within the academic community.

\par\noindent\textbf{Open-Vocabulary 3D Object Detection.} In the field of 3D perception~\cite{chen2023ll3da,Huang2023Segment3D,chen2022d,delitzas2024scenefun3d,yue2024agile3d,peng2023openscene,xu20233difftection,chen2023unit3d,ma2024llms,ge20243d},
the development of open-set 3D object detection (3D-Det) methods has recently demonstrated significant promise across various real-world applications~\cite{luopen}. These methods excel at recognizing unlabeled objects by utilizing a 2D object detection framework trained with ImageNet1K~\cite{deng2009imagenet} to classify outputs from a 3D-Det model~\cite{misra2021end}. This process results in the generation of pseudo labels for the objects identified during training. Subsequently, these pseudo labels are employed to refine and enhance the training of the open-set 3D-Det model~\cite{misra2021end}. Despite the absence of explicit labels for novel objects within the training data, the categories of these novel objects are predetermined and constant, allowing for the creation of category-specific pseudo labels. This defines the method as an open-set yet close-vocabulary model.
More recently, a novel approach has been introduced in~\cite{lu2023open} that addresses the challenges of open-vocabulary 3D object detection (OV-3Ddet). This approach utilizes the CLIP model~\cite{radford2021learning} coupled with a large-scale, pretrained  OV-2Ddet model~\cite{zhou2022detecting} to generate 3D pseudo labels for potentially novel objects. Our work is closer to~\cite{lu2023open}. 
However, rather than generating 3D pseudo labels using pre-existing OV-2Ddet models~\cite{zhou2022detecting}, our novel framework, CoDA~\cite{cao2023coda}, is trained with a limited set of labeled base categories and learn to discover novel objects during training. CoDA is uniquely designed to simultaneously engage in the discovery of novel objects (3D-NOD) and conduct Discovery-Driven Cross-Modal Alignment (DCMA). This collaborative learning strategy empowers our model to efficiently perform end-to-end localization and classification of novel objects.
This work introduces an advanced extension of CoDA, termed CoDAv2. This enhancement primarily extends 3D-NOD by incorporating the newly discovered novel objects into the training scenarios, thereby enriching the presence of novel objects during training and then stimulating the detection of a greater number of novel objects. We designate this enhanced version as 3D Novel Object Discovery with Enrichment (3D-NODE). Furthermore, we improve DCMA by integrating 2D box guidance, which significantly improves discrimination against background regions. This refined approach is termed Discovery-driven Cross-modal Alignment with box guidance (Box-DCMA).

\section{Methodology}\label{sec:methods}
\begin{figure*}[!t]
     \centering
    \begin{overpic}[width=1\textwidth]{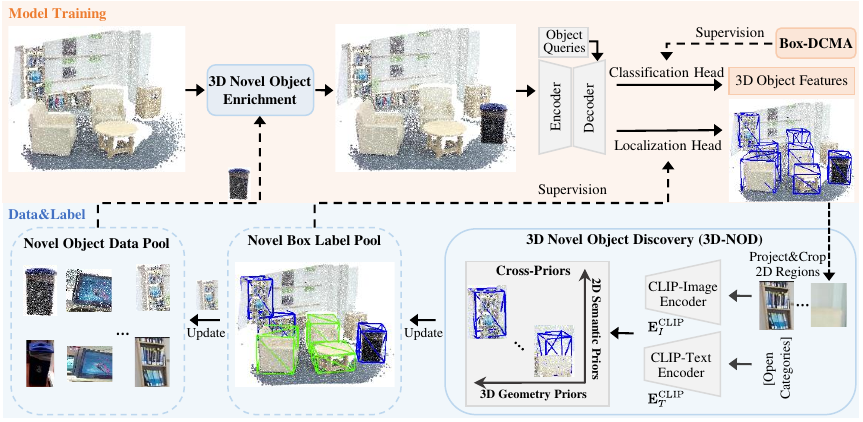}
    \end{overpic}

    \caption{
    Overview of our open-vocabulary 3D object detection system, \textbf{CoDAv2}. Our framework integrates the 3DETR~\cite{misra2021end}, utilizing its Encoder and Decoder networks. In this setup, object queries and point cloud features are processed by the Decoder to refine the query features, which are then directed to the classification and localization heads for 3D objects.
We present a novel method, 3D Novel Object Discovery (\textbf{3D-NOD}), which employs 3D geometric priors from 3D box predictions and 2D open-vocabulary semantic priors from the pre-trained CLIP model, facilitating the discovery of novel objects during the training phase. Building on this, we develop the 3D Novel Object Enrichment approach to extend the 3D-NOD. This approach maintains an online novel object data pool throughout training and augments the training scenes with novel objects sampled from this pool.
The 3D-NOD with Enrichment are collectively termed as \textbf{3D-NODE}.
The discovered novel box labels are also collated in an online novel box label pool to support our novel discovery-driven cross-modal alignment (\textbf{DCMA}), which incorporates class-agnostic distillation alongside class-specific contrastive alignment leveraging these novel boxes. The synergy between 3D-NOD and DCMA promotes effective novel object detection and classification through an end-to-end fashion. Furthermore, we propose to extend the DCMA with 2D box guidance, termed as \textbf{Box-DCMA}, which enhances the ability to distinguish background boxes from foreground, by the predicted 2D boxes from the OV-2DDet model~\cite{liu2024grounding}.
    }

    \label{fig:methods_overview}
\end{figure*} 
\subsection{Framework Overview} \label{sec:method_overview}
The architecture of our open-vocabulary 3D object detection system, \textbf{CoDAv2}, is depicted in~\figref{fig:methods_overview}. This framework integrates the transformer-based point cloud detection model, 3DETR~\cite{misra2021end}, which comprises an `Encoder' and a `Decoder', as illustrated in~\figref{fig:methods_overview}. The model is designed for precise 3D object localization and categorization through its specialized regression and classification heads.
In our approach to facilitate open-vocabulary detection within 3D point cloud environments, we incorporate the pretrained vision-language model, CLIP~\cite{radford2021learning}. This model features dual encoders: CLIP-Image for visual data and CLIP-Text for textual information, which are visible in~\figref{fig:methods_overview}.
To conduct localization for novel objects,
our novel 3D Novel Object Discovery~(\textbf{3D-NOD}) strategy, detailed in~\secref{sec:novel-object-discovery-with-joint-priors}, employs a combination of 3D geometric priors from base categories and 2D semantic insights from CLIP to pinpoint and classify novel objects. 
Further, we extend 3D-NOD with \textbf{3D Novel Object Enrichment}~(\secref{sec:enrichment}),
a strategy that enriches the training scenes with the discovered 
 novel objects from the novel object data pool. This pool is continuously updated online by 3D-NOD.
To conduct classification for novel objects, we design a novel Discovery-Driven Cross-Modal Alignment~(\textbf{DCMA}) module~(\secref{sec:discovery-driven-cross-modality-alignment}). This module is designed to align the features between 3D point clouds and the CLIP's image/text modalities, leveraging the novel objects discovered. 
Then, we propose~\textbf{2D box guidance}~(\secref{sec:box-dcma}) for the extension of DCMA to enhance the classification ability for noise boxes~(background).

\subsection{3D Novel Object Discovery (3D-NOD)} \label{sec:novel-object-discovery-with-joint-priors}
As depicted in~\figref{fig:methods_overview}, our 3D Novel Object Discovery (3D-NOD) strategy facilitates the discovery of novel objects during training by leveraging multi-modality priors from both 3D and 2D domains. In the 3D domain, we utilize geometrical priors from annotated base-category objects to train a class-agnostic box predictor. In the 2D domain, semantic priors from the CLIP model are employed to estimate the likelihood that a 3D object belongs to a novel category. By integrating these dual perspectives, we effectively identify novel object boxes.

\paragraph{Discovery based on 3D Geometry Priors.}
We start with a pool of 3D object box labels $\mathbf{O}_0^{base}$ for base-category objects in the training dataset:
\begin{equation}
\mathbf{O}_0^{base} = \left\{o_j = (l_j, c_j) \mid c_j \in \mathbb{C}^{Seen}\right\},
\label{equ:init_label_set}
\end{equation}
where $\mathbb{C}^{Seen}$ represents the set of base categories with human-annotated labels. Using $\mathbf{O}_0^{base}$, we train an initial 3D object detection model $\mathbf{W}_0^{det}$ by minimizing an object box regression loss inspired by 3DETR~\cite{misra2021end}. This model is trained in a class-agnostic manner, concentrating on box regression and binary objectness prediction, without class-specific classification loss. This approach avoids the limitations of class-specific training, which can hinder the model's ability to identify novel objects, as noted in previous open-vocabulary 2D detection studies~\cite{gu2021open}. Consequently, our modified backbone outputs object boxes along with their 3D localization parameters and objectness scores.
For the $n$-th object query in 3DETR, the model $\mathbf{W}_0^{det}$ predicts its objectness probability $p_n^g$, based on the knowledge from the human-annotated 3D base objects.

\paragraph{Discovery based on 2D CLIP Semantic Priors.} 
To utilize 2D semantic information for 3D object discovery, we first map the 3D object box $l^{3D}_n$ to its corresponding 2D box $l^{2D}_n$ on the 2D image plane using the camera matrix $M$:
\begin{equation}
l^{2D}_n = M \times l^{3D}_n,
\label{equ:3d_to_2d}
\end{equation}
The 2D region defined by $l^{2D}_n$ is then extracted from the image, and its corresponding feature $\mathbf{F}_{I,n}^{Obj}$ is obtained via passing the region through the CLIP image encoder $\mathbf{E}_{I}^\text{CLIP}$.

Given the unknown novel test category list during training, we employ a super category list $T^{super}$, similar to \cite{gupta2019lvis}, to represent a broad range of textual descriptions. These descriptions are encoded using the CLIP text encoder $\mathbf{E}_{T}^\text{CLIP}$, resulting in the text embedding $\mathbf{F}_T^{Super}$. We compute the semantic probability distribution, $\mathbf{P}^{3DObj}_n$, for the 3D object using the corresponding 2D feature:
\begin{equation}
\begin{aligned}
\mathbf{P}^{3DObj}_n &= \left\{ p^{s}_{n,1}, p^{s}_{n,2}, p^{s}_{n,3}, \ldots, p^{s}_{n,C} \right\} \\
&= \mathrm{Softmax}( \mathbf{F}_{I,n}^{Obj} \cdot \mathbf{F}_T^{Super} ),
\label{equ:get_3d_prob}
\end{aligned}
\end{equation}
Here, $C$ represents the number of categories in $T^{super}$, and the dot product operation is denoted by $\cdot$. The distribution $\mathbf{P}^{3DObj}$ serves as the open-vocabulary semantic priors derived from the CLIP model. The object category $c^*$ is determined by the maximum probability in $\mathbf{P}^{3DObj}$, i.e., $c^* = {\operatorname{argmax}}_c \mathbf{P}^{3DObj}$.

To discover novel objects during the $t$-th training epoch, we integrate the objectness score $p_n^g$ from 3D geometry priors with $p^{s}_{n,c^*}$ from 2D semantic priors:
\begin{equation}
\begin{aligned}
\mathbf{O}_t^{disc} = \{ o_j \mid &\forall o_i' \in \mathbf{O}_0^{base}, \mathrm{IoU}_{3D}(o_j, o_i') < 0.25, p_n^g > \theta^g, \\
& p_{n,c^*}^s > \theta^s, c^* \notin \mathbb{C}^{Seen} \},
\end{aligned}
\label{equ:objects_discovered}
\end{equation}
where $\text{IoU}_{3D}$ is the Intersection-over-Union for 3D boxes, and $\theta^s$ and $\theta^g$ are thresholds for semantic and geometry priors, respectively. $\mathbb{C}^{seen}$ is the set of categories seen during training. The newly discovered objects $O_t^{disc}$ are added to the 3D box label pool $\mathbf{O}_t^{novel}$:
\begin{equation}
\mathbf{O}_{t+1}^{novel} = \mathbf{O}_t^{novel} \cup \mathbf{O}_t^{disc}.
\label{equ:update_label}
\end{equation}
This iterative update of the novel box label pool allows the model to improve its ability to localize novel objects. By leveraging the supervision from $O_t^{novel}$, combined with the semantic priors from CLIP and the 3D geometry priors, the model progressively discovers additional novel objects, enhancing its object detection capability for novel objects through the proposed training strategy.

\begin{figure*}[!t]
\centering
\begin{overpic}[width=1\textwidth]{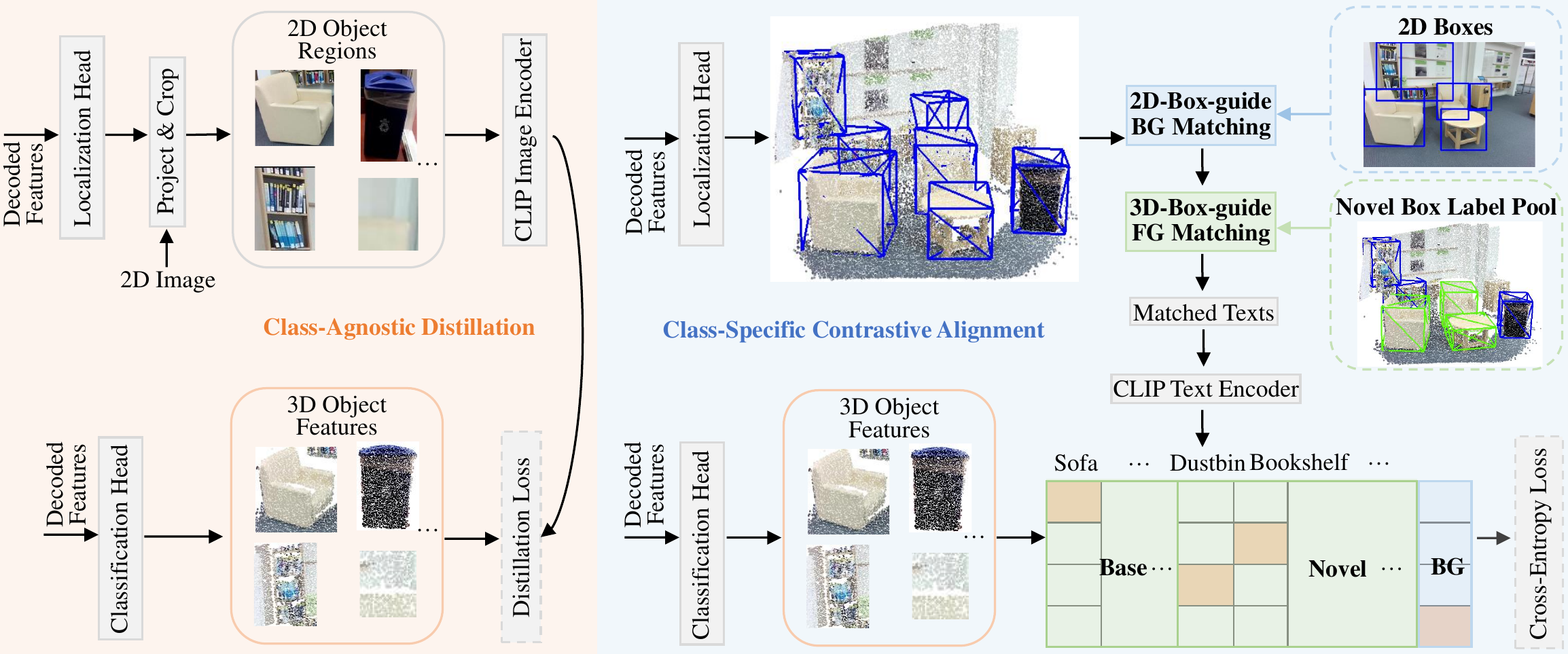}
\end{overpic}
\caption{Illustration of the proposed discovery-driven cross-modal alignment with box guidance (\textbf{Box-DCMA}) within our open-vocabulary detection framework (\textbf{CoDAv2}). The Box-DCMA consists of two primary components: the class-agnostic distillation (left panel) and the class-specific contrastive feature alignment (right panel). Initially, predicted 3D boxes from the detection head are projected to obtain 2D image object regions. These regions are then encoded by the CLIP image encoder to produce their corresponding 2D object features. Next, these 2D features derived from CLIP and the 3D point cloud object features are fed into the class-agnostic distillation module to achieve feature alignment.
Besides, the dynamically updated novel box label pool that is driven by 3D-NOD, is leveraged to align the predicted 3D foreground object boxes with foreground category texts through the 3D-Box-guide FG Matching during training. Background regions are aligned with background category texts using 2D-Box-guide BG Matching. We perform contrastive alignment for the matched boxes to learn more discriminative 3D object features. These enhanced 3D object features subsequently aid in the accurate prediction of novel objects.}
    \label{fig:alignment}
\end{figure*}

\subsection{3D Novel Object Enrichment} \label{sec:enrichment}
As shown in~\figref{fig:methods_overview}, our 3D Novel Object Enrichment maintains an online novel object data pool during training and augments the training scenes with novel objects from the pool. By broadening the distribution of novel objects, this method enhances the detection capabilities for 3D novel objects.
Specifically, after discovering boxes of novel objects during training in 3D-NOD, we can extract the corresponding point cloud objects in the 3D boxes by:
\begin{equation}
\mathbf{PC}_t^{disc} = \mathrm{Extract}(\mathbf{O}_t^{disc}, \mathbf{PC}_t^{scene}),
\label{equ:extract_pc}
\end{equation}
where $\mathbf{O}_t^{disc}$ represents the discovered novel object boxes in the $t$-th training epoch, as shown in~\equref{equ:objects_discovered}. $\mathbf{PC}_t^{scene}$ is the input point cloud scene. $\mathbf{PC}_t^{disc}$ denotes the novel object point clouds within the discovered 3D boxes $\mathbf{O}_t^{disc}$ in the point cloud scene $\mathbf{PC}_t^{scene}$.
Furthermore, we can obtain the corresponding 2D object boxes of the discovered novel objects using the camera matrix $M$ as follows:
\begin{equation}
\mathbf{B}_t^{disc} = M \times \mathbf{O}_t^{disc},
\label{equ:2d_regions}
\end{equation}
Then, we can obtain the 2D object regions $\mathbf{I}_t^{disc}$ by cropping the 2D images with $\mathbf{B}_t^{disc}$.
Furthermore, the discovered novel object point clouds and 2D regions are updated into the maintained novel object data pool as follows:
\begin{equation}
\mathbf{D}_{t+1}^{novel} = \mathbf{D}_t^{novel} \cup (\mathbf{PC}_t^{disc}, \mathbf{I}_t^{disc}),
\label{equ:update_pool}
\end{equation}
where $\mathbf{D}_{t+1}^{novel}$ represents the novel object data pool in the $(t+1)$-th training epoch, which includes both the data pairs of 3D and 2D objects.
Then, during the $(t+1)$-th training epoch, we will randomly sample novel objects from the novel object data pool $\mathbf{D}_{t+1}^{novel}$ and insert them into the input point cloud scene $\mathbf{PC}_{t+1}^{scene}$, as shown in~\figref{fig:methods_overview}. 

\updateblue{The details of the object insertion strategy are as follows:
We first randomly select positions within the room space to place objects. To prevent inserted objects from embedding into walls or overlapping with existing objects, we perform an occlusion check. Specifically, we count the number of points, $N_{\text{check}}$, inside the bounding box volume $\mathbf{O}_t^{disc'}$ of the inserted object. If $N_{\text{check}} > J$, where $J$ is an empirically determined threshold~(set to 1000 in our experiments), we discard the current position and randomly select a new one. 
We repeat this process until $N_{\text{check}} \leq J$,
ensuring small overlaps between the inserted object and the training scene.}

\updateblue{The primary goal of 3D-NODE is to enrich the training dataset with discovered novel objects to improve the open-vocabulary 3D detection~(OV-3DDet), rather than to generate strictly accurate indoor scenes.
Thus, we did not incorporate additional layout priors to prevent objects from floating.
Our easy-to-apply and efficient insertion strategy, as described above, sufficiently serves our goal,
and effectively improves the performance~(7.53 vs. 6.71 regarding $\text{AP}_{Novel}$, 37.60 vs. 33.66 regarding $\text{AR}_{Novel}$), as shown in~\tabref{tab:main-ablation}~(`3D-NODE + DCMA' vs. `3D-NOD + DCMA').
A similar observation has been reported in prior 2D instance segmentation work~\cite{ghiasi2021simple}, which stated: ``the simple mechanism of pasting objects randomly is good enough and can provide solid gains on top of strong baselines.''
Our design has also been proven to work effectively for OV-3DDet.}

\updateblue{
In summary, our 3D Novel Object Enrichment
}
enriches the distribution of novel objects in the training set, leading to more effective learning for novel object detection.

\subsection{Discovery-Driven Cross-Modal Alignment (DCMA)} \label{sec:discovery-driven-cross-modality-alignment}
To gain open-vocabulary capabilities from the CLIP model, we introduce a discovery-driven cross-modal alignment (DCMA) module. This module aligns 3D point cloud object features with 2D-image feature, and with CLIP-text features, leveraging the discovered novel object boxes $\mathbf{O}^{novel}$. The DCMA module involves two primary cross-modal feature alignment strategies, i.e., class-agnostic distillation and class-specific contrastive alignment.

\paragraph{Class-agnostic Distillation.} 
We employ a class-agnostic distillation scheme to align 3D point cloud object features with 2D image object features, facilitating the transfer of open-vocabulary knowledge from the CLIP model. As for the $n$-th object query, the updated 3D object feature $\mathbf{F}_n^{3dObj}$ is obtained at the final decoder layer. Using this, the model predicts the 3D box parameters $l^{3D}_n$ and derives the corresponding 2D box $l^{2D}_n$ via~\equref{equ:3d_to_2d}. The projected 2D object features $\mathbf{F}^{2DObj}_n$ are then obtained by inputting the cropped 2D region based on $l^{2D}$ into CLIP. To minimize the distance between 3D and 2D object feature spaces, we use a class-agnostic L1 distillation loss, defined as:
\begin{equation}
\mathcal{L}_{distill}^{3D\leftrightarrow 2D} = \sum_{n=1}^N |\mathbf{F}^{3DObj}_n - \mathbf{F}^{2DObj}_n|,
\label{equ:distill}
\end{equation}
where $N$ is the number of object queries. As demonstrated in~\figref{fig:alignment}, even when the boxes cover background regions, this distillation process can effectively narrow the gap between modalities, enhancing general feature alignment across different scenes. Notably, this method does not require category labels for ground-truth boxes, making $\mathcal{L}_{distill}^{3D\leftrightarrow 2D}$ independent of category annotations.

\paragraph{Discovery-driven Class-specific Contrastive Alignment.}
In \secref{sec:novel-object-discovery-with-joint-priors},
we modified the 3DETR detection backbone by excluding the fixed-category classification head, and generating 3D point cloud object features from the updated object queries.
For the $n$-th object query, we obtain the 3D object feature $\mathbf{F}^{3DObj}_n$ at the last decoder layer. To ensure these features are discriminative for both novel and base categories within the superset $T^\text{super}$, we conduct the class-specific alignment with CLIP text features.

We compute the normalized similarities using the dot product operation between the 3D object feature $\mathbf{F}^{3DObj}_n$ and the text feature $\mathbf{F}_T^{Super}$ as follows:
\begin{equation}
\mathbf{S}_n = \mathrm{Softmax}(\mathbf{F}^{3DObj}_n \cdot \mathbf{F}_T^{Super}),
\end{equation}
We then introduce a contrastive loss to supervise the learning of 3D object features. After measuring the similarity between the 3D object point cloud features and the text features, we identify the most similar pairs as ground-truth labels using Bipartite Matching from~\cite{misra2021end}. This ensures class-specific alignment is applied to foreground novel objects. Upon the matching with the box label pool $\mathbf{O}^{label}$, we derive the category label from the CLIP model by finding the maximum similarity score, forming a one-hot label vector $\mathbf{h}_n$ for $\mathbf{F}^{3DObj}_n$. This process is illustrated in~\figref{fig:alignment}. Utilizing discovered foreground novel object boxes, we then define a discovery-driven cross-modal contrastive loss:
\begin{equation}
\mathcal{L}_{contra}^{3D \leftrightarrow Text} = \sum_{n=1}^N \mathbf{1}(\mathbf{F}^{3DObj}_n, \mathbf{O}^{label}) \cdot \mathrm{CE}(\mathbf{S}_n, \mathbf{h}_n),
\label{equ:contrastive}
\end{equation}
where $N$ is the number of object queries; $\mathbf{1}(\cdot)$ is an indicator function which returns 1 for a match between the query and the box label pool otherwise 0 for a non-match; $\mathrm{CE}(\cdot)$ denotes the cross-entropy loss. This strategy ensures the involvement of novel object queries in the cross-modal alignment. It improves feature alignment learning and makes 3D features of novel objects more discriminative, thereby enhancing the model capability in the discovery of novel objects. Consequently, the integrated process of 3D novel object discovery and cross-modal feature alignment synergistically advances the objectives of localization and classification in open-vocabulary 3D object detection.

\subsection{Box Guidance for DCMA} \label{sec:box-dcma}
Considering that the CLIP model~\cite{radford2021learning} tends to perform unsatisfactorily on background regions, our framework, which derives the open-vocabulary classification capability from CLIP, may inadvertently classify background regions as object regions. To enhance the discrimination ability on background areas, we extend our DCMA module with 2D box guidance, as illustrated in `2D-Box-guide BG Matching' in~\figref{fig:alignment}.
Specifically, we initially detect the 2D bounding boxes of objects in the training set images using an OV-2DDet model~\cite{liu2024grounding}, defined as:
\begin{equation}
\mathbf{B^{refer}_{2D}} = \mathrm{H}(\mathbf{I}),
\label{equ:get_ov_2d_boxes}
\end{equation}
where $\mathbf{I}$ represents the 2D image and $\mathrm{H}$ denotes the OV-2DDet model~\cite{liu2024grounding}. Here, $\mathbf{B^{refer}_{2D}}$ indicates the detected 2D reference boxes, which serve as references for the subsequent occlusion check.
We then project the predicted 3D boxes, $\mathbf{B^{pred}_{3D}}$, onto the 2D image plane using the camera matrix, and then obtain the 2D boxes as follows:
\begin{equation}
\mathbf{B^{pred}_{2D}} = M \times \mathbf{B^{pred}_{3D}},
\label{equ:3d_to_2d_check}
\end{equation}
where $\mathbf{B^{pred}_{2D}}$ represents the 2D boxes projected from the predicted 3D boxes.
Following this, we compute the Intersection over Union (IoU) between every pair of projected 2D boxes and the reference boxes detected by the OV-2DDet model to perform an occlusion check:

\begin{equation}
\begin{aligned}
\mathbf{V_{iou}} = \left\{
\mathrm{IOU}(\mathbf{b^{pred}_{2D}}, \mathbf{b^{refer}_{2D}}) \mid \right. &\mathbf{b^{pred}_{2D}} \in \mathbf{B^{pred}_{2D}},  \\
&\left. \mathbf{b^{refer}_{2D}} \in \mathbf{B^{refer}_{2D}}  \right\}
\end{aligned}
\label{equ:oc_check}
\end{equation}
where the set $\mathbf{V_{iou}}$ contains the IoU values for every possible pairing of bounding boxes from the predicted and reference sets, $\mathbf{B^{pred}_{2D}}$ and $\mathbf{B^{refer}_{2D}}$, respectively.
If the maximum IoU in set~$\mathbf{V_{iou}}$ is lower than $k$, we assign the `background' label to the corresponding predicted 3D boxes. 
The following alignment for
the background category is incorporated
into the same process for other foreground categories
in DCMA as introduced in~\secref{sec:discovery-driven-cross-modality-alignment}.

\subsection{Overall Framework Optimization}

For the optimization of the entire CoDAv2 framework, as we consider 3DETR~\cite{misra2021end} as our base model, we thus first adopt the loss function in~\cite{misra2021end} defined as follows:
\begin{multline}
\mathcal{L'}_{\mathrm{\OURS}} =   -\lambda_{ac} \ba_c^\intercal \log \hba_c + \lambda_{ar} \|\hba_r - \ba_r\|_{\mathrm{huber}} + \lambda_{d} \|\hbd - \bd\|_{1} \\
+ \lambda_{c} \|\hbc - \bc\|_{1}  
 -\lambda_{s} \bsd_c^\intercal \log \hbsd_c,
\end{multline}
where includes the angle loss~($-\lambda_{ac} \ba_c^\intercal \log \hba_c + \lambda_{ar} \|\hba_r - \ba_r\|_{\mathrm{huber}}$), the center loss~($\lambda_{c} \|\hbc - \bc\|_{1}$), the size loss~($\lambda_{d} \|\hbd - \bd\|_{1}$), and the semantic classification loss~($-\lambda_{s} \bsd_c^\intercal \log \hbsd_c$).
Particularly for the semantic classification loss, we have configured it as a binary classification task to facilitate class-agnostic categorization. The hyperparameters of the loss weights are set following the 3DETR~\cite{misra2021end}. Besides the 3DETR loss function, we additionally integrate our alignment losses within DCMA as described in~\secref{sec:discovery-driven-cross-modality-alignment},
including the class-agnostic loss $\mathcal{L}_{distill}^{3D\leftrightarrow 2D}$ in~\equref{equ:distill} and class-specific loss $\mathcal{L}_{contra}^{3D \leftrightarrow Text}$ in~\equref{equ:contrastive}. Then, the overall optimization loss function $\mathcal{L}_{final}$ can be formulated as:
\begin{equation}
    \mathcal{L}_{final} = \mathcal{L}_{distill}^{3D\leftrightarrow 2D} + \mathcal{L}_{contra}^{3D \leftrightarrow Text} + \mathcal{L'}_{\mathrm{\OURS}}.
\end{equation}

\section{Experiments}\label{sec:experiments}
\subsection{Experimental Setup}
\textbf{Datasets and Settings.} Our experiments are conducted on two challenging indoor 3D object detection datasets, i.e., SUN-RGBD~\cite{song2015sun} and ScanNetv2~\cite{dai2017scannet}. SUN-RGBD consists of 5,000 training samples spanning 46 object categories, while ScanNetv2 contains 12,000 training samples~\cite{rozenberszki2022language}.
For both training and evaluation, we adopt a strategy similar to those used in 2D open-vocabulary object detection works~\cite{gu2021open, zareian2021open}. Specifically, it categorizes the object classes into base (seen) and novel categories based on sample counts in each category. In SUN-RGBD, the top 10 categories with the most training samples are classified as base, and the remaining 36 as novel. In ScanNetv2, the top 10 categories are considered as base, while the other 50 are treated as novel.
Point cloud scenes are generated from depth maps using the method described in~\cite{lu2023open}. The evaluation metric used is mean Average Precision (mAP) at an IoU threshold of 0.25, denoted as $\text{AP}_{25}$, following~\cite{misra2021end}.

\paragraph{Implementation Details.}
We set the batch size to 8, and we set the number of object queries to 128 for both SUN-RGBD and ScanNetv2.
Initially, we train a base 3DETR model for 1080 epochs utilizing only the class-agnostic distillation. 
Then, we incorporate the extended 3D novel object
discovery with enrichment~(3D-NODE) and discovery-driven cross-modal
feature alignment with box guidance~(Box-DCMA), and the model continues to be trained for 400 epochs on SUN-RGBD and 600 epochs on ScanNetv2.
The novel box label pool is updated every 50 epochs. As for the feature alignment in the proposed Box-DCMA, considering the limited number of ground truth (GT) and discovered boxes in each scene, we use a larger number of object queries to facilitate the contrastive learning. 
{For the 128 queries, aside from those matched
by foreground matching and background matching, we select an additional 32 queries for alignment with the CLIP model.}
The hyper-parameters used during training follow the default 3DETR configuration as specified in~\cite{misra2021end,cao2023coda}.

\subsection{Model Analysis}
In this section, we conduct a comprehensive ablation study to evaluate individual contributions of each component within our open-vocabulary detection framework CoDAv2. To systematically assess the effectiveness of our proposed designs, we incrementally integrate them into the backbone detection architecture. Our analysis proceeds in the following order: class-agnostic distillation~(\secref{exp:cad}),
3D-NOD~(\secref{exp:3d-nod}), DCMA~(\secref{exp:dcma}), 3D-NODE~(\secref{exp:3d-node}), and Box-DCMA~(\secref{exp:box-dcma}).
This evaluation not only highlights the performance enhancements each component brings but also delves into specific design choices and hyper-parameter configurations.
In~\secref{exp:cad}, we explore the effect of class-agnostic distillation.
Moving to~\secref{exp:3d-nod}, we examine the effect of
3D-NOD and the thresholds for geometry and semantic priors in 3D-NOD.
In~\secref{exp:dcma}, our analysis studies the effect of DCMA, covering various aspects including class-agnostic vs. class-specific distillation,
the effect of different matching methods in DCMA, and the impact of collaborative learning between 3D-NOD and DCMA.
\secref{exp:3d-node} focuses on investigating the effect of 3D-NODE, and the number of novel objects inserted into the training 3D scenes in the enrichment strategy.
\secref{exp:box-dcma} investigates the effect of Box-DCMA, different matching methods for the `background' category in Box-DCMA, the IoU threshold in Box-DCMA.
Finally, in~\secref{sec:analysis_method},
we conduct an ablation study for different numbers of test categories in the evaluation.
The models are tested across three subsets of categories: novel categories, base categories, and all categories. The performances on these subsets are indicated as $\text{AP}_{Novel}$, $\text{AP}_{Base}$, and $\text{AP}_{Mean}$, respectively. Evaluation metrics, including mean Average Precision (mAP) and recall ($\text{AR}$), are computed at an IoU threshold of 0.25. Extensive ablation studies are performed using the SUN-RGBD dataset.

\begin{table*}[!t]
    \centering
    \caption{Ablation study to validate the effectiveness of the designed 3D novel object discovery~(3D-NOD), discovery-driven cross-modality alignment~(DCMA),
    3D novel object discovery with enrichment~(3D-NODE), and discovery-driven cross-modality alignment with box guidance~(Box-DCMA)
    in our proposed 3D OV detection framework CoDAv2. `Distillation' represents our base model, a 3DETR backbone that is trained with only class-agnostic distillation. As for `3D-NOD+Distillation \& PlainA',
we introduce plain point-text alignment, which aligns the category list of base category texts without utilizing the discovered boxes by 3D-NOD.}

    \label{tab:main-ablation}    
    \newcommand{\tf}[1]{\textbf{#1}}
    \resizebox{0.9\textwidth}{!}{
    \begin{tabular}{lcccccc}
    \toprule
Methods&  $\text{AP}_{Novel}$ & $\text{AP}_{Base}$ & $\text{AP}_{Mean}$ & $\text{AR}_{Novel}$ & $\text{AR}_{Base}$ & $\text{AR}_{Mean}$ \\
\midrule
3D-CLIP~\cite{radford2021learning} & 3.61& 30.56& 9.47& 21.47& 63.74& 30.66 \\
Distillation& 3.28& 33.00& 9.74& 19.87& 64.31& 29.53  \\
3D-NOD + Distillation& 5.48& 32.07& 11.26& 33.45&	66.03& 40.53 \\
3D-NOD + Distillation \& PlainA& 0.85& 35.23& 8.32&	34.44& 66.11& 41.33 \\
3D-NOD + DCMA & 6.71& 38.72& 13.66& 33.66& 66.42& 40.78 \\
3D-NODE + DCMA & 7.53& 41.06& 14.82& 37.60& 69.39& 44.51 \\
3D-NODE + Box-DCMA & \textbf{9.17}& \textbf{42.04}& \textbf{16.31}& \textbf{43.16}& \textbf{71.64}& \textbf{49.35} \\
    \bottomrule
    \end{tabular}
    }
\end{table*}

\begin{figure}[!t]
\centering
    \begin{overpic}[width=0.93\linewidth]{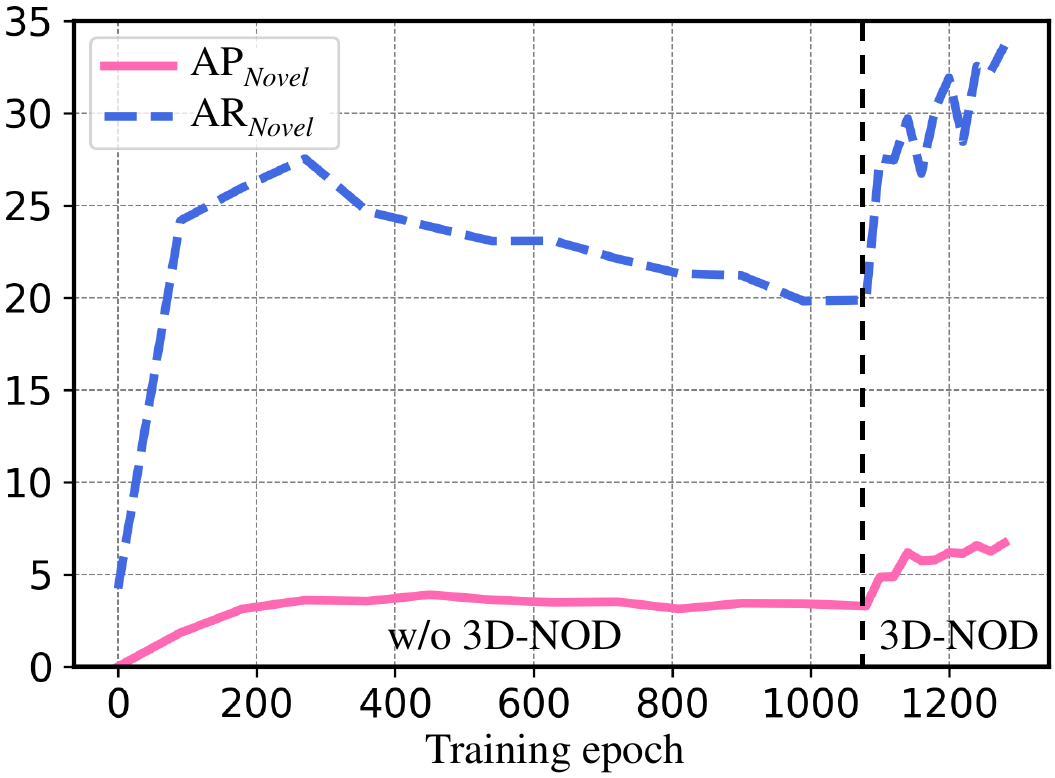}
    \end{overpic}
    \caption{Effect of 3D Novel Object Discovery~(3D-NOD) during training. The pink curve represents $\text{AP}_{novel}$, and the blue curve represents $\text{AR}_{novel}$. The left side of the black dashed line shows the training process of our base model `Distillation', trained for 1080 epochs. The right side represents an additional 200 epochs of training with the insertion of 3D-NOD. The implementation of 3D-NOD results in significant increases in both $\text{AP}_{novel}$ and $\text{AR}_{novel}$, effectively overcoming the limitations of base category annotations.}

\label{fig:recall_curve}
\end{figure}

\subsubsection{Analysis of Class-agnostic Distillation}
\label{exp:cad}
\vspace{2mm}
We train a 3DETR model exclusively with class-agnostic distillation, which serves as our base model, denoted as `Distillation' in~\tabref{tab:main-ablation}. 
Besides, to analyze the impact of class-agnostic distillation on discovery-driven alignment, we also build a baseline model by
training a class-agnostic 3DETR detection model that accommodates basic 3D localization of novel objects. These 3D bounding boxes are projected onto 2D color images to extract their corresponding 2D object regions. By processing these regions through the CLIP model, we enable open-vocabulary classification, thereby forming an open-vocabulary 3D detection framework that requires both point cloud and image inputs. This baseline approach is referred to as `3D-CLIP' in~\tabref{tab:main-ablation}.
The results in the table reveal that, when evaluated using only point cloud inputs, the `Distillation' model achieves performance comparable to the `3D-CLIP' model in terms of $\text{AP}_{Novel}$ (3.28 vs. 3.61), suggesting that class-agnostic distillation endows 3DETR with preliminary open-vocabulary detection capabilities.
Furthermore, since class-agnostic distillation aligns 3D point cloud object features with 2D image features wherever the predicted 3D bounding boxes cover, and because base categories have more training samples, distillation on these categories proves more effective. This effectiveness is reflected in the higher $\text{AP}_{Base}$ performance (33.00 vs. 30.56).

\subsubsection{Analysis of 3D Novel Object Discovery~(3D-NOD)}
\label{exp:3d-nod}
\vspace{2mm}
\paragraph{Effect of 3D Novel Object Discovery~(3D-NOD)}.
We enhance the base model `Distillation' by incorporating our 3D novel object discovery (3D-NOD) strategy, resulting in the `3D-NOD + Distillation' model as shown in~\tabref{tab:main-ablation}. The `3D-NOD + Distillation' model achieves a higher recall than both `Distillation' and `3D-CLIP' (33.45 vs. 19.87/21.47), indicating that 3D-NOD discovers more novel objects during training and thus improves $\text{AP}_{Novel}$ (5.48 vs. 3.28/3.61).
To further investigate the impact of 3D-NOD during training, we track the model's performance at intermediate checkpoints. \figref{fig:recall_curve} shows $\text{AP}_{novel}$ (pink curve) and $\text{AR}_{novel}$ (blue curve) over the training period. The base model `Distillation', trained for 1080 epochs, corresponds to the left side of the black dashed line. After applying the 3D-NOD, training continues for an additional 200 epochs, represented on the right side of the dashed line. Without 3D-NOD, the $\text{AP}_{novel}$ of the base model plateaus around 1000 epochs, and $\text{AR}_{novel}$ starts to decline due to the category forgetting phenomenon. However, with 3D-NOD, significant increases in both $\text{AP}_{novel}$ and $\text{AR}_{novel}$ are observed, validating the effectiveness of the 3D-NOD approach.

\paragraph{Effect of Geometry and Semantic Priors in 3D-NOD}.
The 3D-NOD strategy leverages 2D semantic priors and 3D geometry priors, each requiring a threshold. We evaluate the sensitivity of these thresholds using the `3D-NOD + DCMA' model in~\tabref{tab:main-ablation}, with results shown in~\tabref{tab:sensitivity}. The first row (`0.0') represents the base model without our 3D-NOD. Our experiments illustrate that 3D-NOD consistently improves the OV detection performance across a wide range of threshold settings, indicating the robustness of model. As evidenced in~\tabref{tab:sensitivity}, all models trained with different thresholds outperform the baseline (`0.0\&0.0') by at least 70\%, demonstrating the clear benefit of 3D-NOD. The threshold setting of 0.3 yields the best performance, facilitating the discovery of more novel objects.

\begin{table}[!t]
    \centering
    \caption{Comparison between class-specific contrastive loss and class-agnostic distillation in DCMA.}
    \label{tab:contrast&distillation}    
    \newcommand{\tf}[1]{\textbf{#1}}
    \resizebox{1\linewidth}{!}{
    \LARGE
    \begin{tabular}{lcccccc}
    \toprule[1.6pt]
Methods&  $\text{AP}_{Novel}$ & $\text{AP}_{Base}$ & $\text{AP}_{Mean}$ & $\text{AR}_{Novel}$ & $\text{AR}_{Base}$ & $\text{AR}_{Mean}$ \\
\midrule[1pt]
Class-specific 
& 5.35& 34.58& 11.70&	37.93& 66.93& 44.24 \\
Class-agnostic 
& \textbf{6.71}& \textbf{38.72}& \textbf{13.66}& 33.66& 66.42& 40.78 \\

    \bottomrule[1.6pt]
    \end{tabular}
    }
\end{table}

\begin{table}[!t]
    \centering
    \caption{Ablation study  to evaluate the impact of the collaborative learning between DCMA and 3D-NOD. In comparison to `3D-CLIP + 3D-NOD', the joint utilization of 3D-NOD and DCMA significantly improves the open-vocabulary 3D detection performance.}
    \label{tab:dcma-on-3dnod}    
    \newcommand{\tf}[1]{\textbf{#1}}
    \resizebox{\linewidth}{!}{
    \Huge
    \begin{tabular}{lcccccc}
    \toprule[2.3pt]
Methods&  $\text{AP}_{Novel}$ & $\text{AP}_{Base}$ & $\text{AP}_{Mean}$ & $\text{AR}_{Novel}$ & $\text{AR}_{Base}$ & $\text{AR}_{Mean}$ \\
\midrule[1.5pt]
3D-CLIP~\cite{radford2021learning} & 3.61& 30.56& 9.47& 21.47& 63.74& 30.66 \\
3D-CLIP + 3D-NOD& 5.30& 26.08& 9.82& 32.72& 64.43& 39.62  \\
3D-NOD + DCMA & \textbf{6.71}& \textbf{38.72}& \textbf{13.66}& 33.66& 66.42& 40.78 \\

    \bottomrule[2.3pt]
    \end{tabular}
    }
\end{table}

\begin{table}[!t]
    \centering
    \caption{Ablation study on the thresholds used for the 3D geometry priors and open-vocabulary 2D semantic priors in the 3D Novel Object Discovery (3D-NOD) strategy. Disabling these priors results in significant performance drops (the first row). The model gains the best $\text{AP}_{Novel}$ performance with a threshold of 0.3.}
            \label{tab:sensitivity}
    \newcommand{\tf}[1]{\textbf{#1}}
    \resizebox{1\columnwidth}{!}{
    \Huge
    \begin{tabular}{cccccccc}
    \toprule[2.5pt]
 Semantic & Geometry& $\text{AP}_{Novel}$ & $\text{AP}_{Base}$ & $\text{AP}_{Mean}$ & $\text{AR}_{Novel}$ & $\text{AR}_{Base}$ & $\text{AR}_{Mean}$ \\
    \midrule[1.5pt]
0.0 & 0.0 & 3.28& 33.00 & 9.74& 19.87& 64.31& 29.53 \\
0.3 & 0.3& \textbf{6.71}& 38.72& \textbf{13.66}& 33.66& 66.42& 40.78 \\
0.3 & 0.5& 6.35& \textbf{39.57}& 13.57& 31.93& 66.91& 39.53 \\
0.5 & 0.3& 5.70 &	38.61& 12.85&	27.05& 63.61& 35.00
 \\
0.5 & 0.5& 5.70& 39.25& 13.00& 28.27& 65.04& 36.26
 \\
    \bottomrule[2.5pt]
    \end{tabular}}
\end{table}

\subsubsection{Analysis of Discovery-driven Cross-modal Alignment~(DCMA)}
\label{exp:dcma}
\vspace{2mm}
\paragraph{Effect of DCMA in CoDAv2}.
We investigate the influence of various alignment strategies on the `3D-NOD + Distillation' model. Initially, we integrate a plain text-point alignment approach, denoted as `3D-NOD + Distillation \& PlainA'. This method aligns the category list of base category texts without leveraging the discovered boxes from 3D-NOD. According to~\tabref{tab:main-ablation}, the $\text{AP}_{Base}$ for `3D-NOD + Distillation \& PlainA' is higher than that of `3D-NOD + Distillation' (35.23 vs. 32.07), indicating that base category texts enhance cross-modal feature alignment. Additionally, 3D-NOD continues to improve $\text{AR}_{Novel}$, showing that it still aids in discovering novel objects in this context. However, the $\text{AP}_{Novel}$ is lower at 0.85, suggesting that without utilizing box discovery from 3D-NOD, the model's discriminative power on novel categories diminishes.
Conversely, the `3D-NOD + DCMA' model in~\tabref{tab:main-ablation} that applies our discovery-driven class-specific cross-modal alignment method, achieves superior performance on both $\text{AP}_{Novel}$ and $\text{AP}_{Base}$. This indicates that, compared to plain alignment, our discovery-driven alignment approach yields more discriminative object features by using an expanded category vocabulary facilitated by 3D-NOD.

\paragraph{Class-agnostic Distillation vs. Class-specific Distillation}.
Class-specific contrastive learning~\cite{lu2023open} promotes similarity among features of the same category. However, during the discovery process, the discovered boxes might include background regions. Since CLIP cannot classify backgrounds, high-scoring boxes on background regions may be incorrectly assigned to categories, causing significant feature misalignment with class-specific contrastive learning. In contrast, our class-agnostic distillation aligns 3D object query features with their corresponding projected 2D image features, covering both foreground and background 3D object boxes. Thus, distillation effectively handles both foreground and background simultaneously.
To demonstrate the advantage of our approach over class-specific contrastive learning, we replace class-agnostic distillation with class-specific contrastive learning in our method. As shown in~\tabref{tab:contrast&distillation}, while the contrastive loss increases the recall by classifying more boxes as foreground, it also misclassifies more background boxes as foreground objects. This results in a lower overall mAP compared to our class-agnostic distillation approach.

\paragraph{Effect of Different Matching Strategies in DCMA}.
In discovery-driven alignment, matching predicted boxes with those in the discovered box label pool is essential for constructing the contrastive alignment loss. These matched predicted boxes are used in the contrastive feature alignment, as illustrated in~\figref{fig:alignment}. We compare different matching strategies, specifically 3DIoU match and Bipartite match. For 3DIoU match, predicted 3D boxes are activated if their 3DIoU value exceeds 0.25 with any box in the label pool.
As detailed in~\tabref{tab:match}, {`DCMA w/o Matching' represents the model trained without matching and the alignment between the 3D and text modalities, which corresponds to `3D-NOD + Distillation' in~\tabref{tab:main-ablation}.} The $\text{AP}_{Novel}$ and $\text{AP}_{Base}$ for 3DIoU matching are lower than those for `DCMA w/ Bipartite' and comparable to `DCMA w/o Matching', indicating the ineffectiveness of the 3DIoU matching method. This is primarily due to the higher density of indoor scenes compared to outdoor scenes. Consequently, a shift in the predicted box can result in covering different objects. Moreover, 3DIoU matching may assign multiple predicted boxes to a single 3D ground-truth box if their 3DIoU values exceed 0.25, causing semantic confusion and invalidating the alignment. Therefore, Bipartite matching is preferred over 3DIoU matching in our discovery-driven alignment.

\begin{table}[!t]
    \centering
    \caption{Comparison of different matching strategies in DCMA. `DCMA w/o Matching' represents the model trained without matching and without aligning 3D and text modality, i.e., the `3D-NOD + Distillation' in~\tabref{tab:main-ablation}.}
    \label{tab:match}    
    \newcommand{\tf}[1]{\textbf{#1}}
    \resizebox{\linewidth}{!}{
    \Huge
    \begin{tabular}{lcccccc}
    \toprule[2.5pt]
Methods&  $\text{AP}_{Novel}$ & $\text{AP}_{Base}$ & $\text{AP}_{Mean}$ & $\text{AR}_{Novel}$ & $\text{AR}_{Base}$ & $\text{AR}_{Mean}$ \\
    \midrule[1.5pt]
DCMA w/o Matching& 5.48& 32.07& 11.26& 33.45&	66.03& 40.53 \\
DCMA w/ 3DIoU & 5.68& 31.29& 11.25& 33.08& 67.35& 40.53 \\
DCMA w/ Bipartite & \textbf{6.71}& \textbf{38.72}& \textbf{13.66}& 33.66& 66.42& 40.78 \\
    \bottomrule[2.5pt]
    \end{tabular}
    }
    \vspace{3pt}
\end{table}

\paragraph{Effect of the Collaborative Learning of 3D-NOD and DCMA}.
As indicated in~\tabref{tab:main-ablation}, we conduct an ablation study to compare our method with the alignment method not driven by 3D-NOD ({i.e.}, `3D-NOD + DCMA' vs. `3D-NOD + Distillation \& PlainA').
It shows that our 3D-NOD method enhances cross-modal alignment, leading to an improved performance in novel category classification.
To examine the effect of DCMA on 3D-NOD, we integrate our 3D-NOD method into the `3D-CLIP' pipeline without using DCMA, denoted as `3D-CLIP + 3D-NOD' in~\tabref{tab:dcma-on-3dnod}. As shown in~\tabref{tab:dcma-on-3dnod}, `3D-NOD + DCMA' surpasses `3D-CLIP + 3D-NOD' in terms of $\text{AP}$ and $\text{AR}$ for both base and novel categories. This demonstrates that DCMA effectively helps to learn more discriminative feature representations of object query embeddings, thereby enhancing localization capability.
Overall, these findings indicate that the joint training of 3D-NOD and DCMA can improve the performance by 48\% on $\text{AP}_{Base}$ and 26\% on $\text{AP}_{Novel}$.

\begin{figure}[!t]
\centering
    \begin{overpic}[width=1\linewidth]{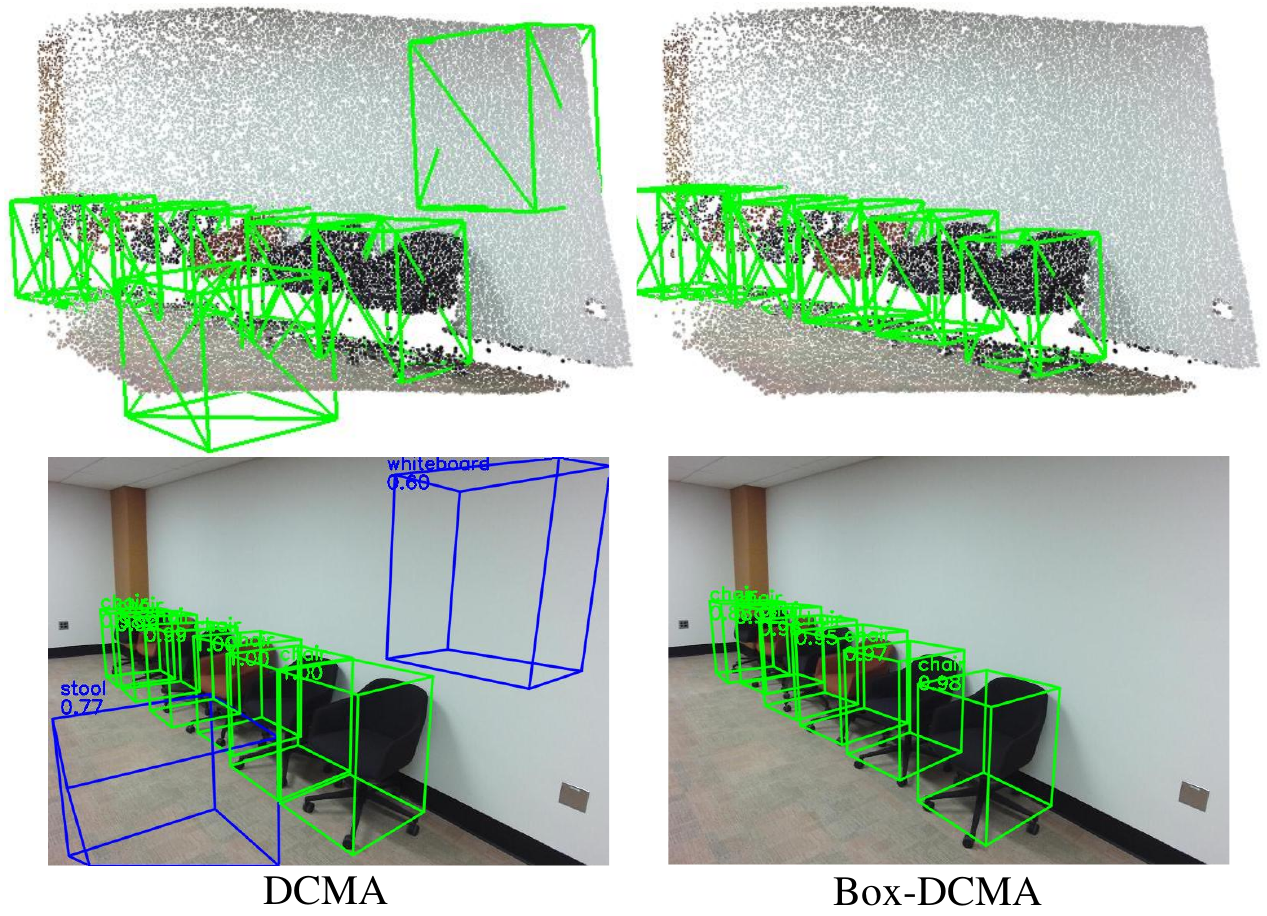}
    \end{overpic}
    \caption{Qualitative illustration of the impact of the box guidance on noisy~(background) boxes estimation.
   Compared with `DCAM', `Box-DCMA' detects less background boxes, benefiting from the proposed box guidance. }

\label{fig:box_dcma}
\end{figure}

\begin{table}[!t]
    \centering
    \caption{Comparison of different matching methods in the `2D-Box-guide BG Matching' of Box-DCMA.}
    \label{tab:assign_DCMA}    
    \newcommand{\tf}[1]{\textbf{#1}}
    \resizebox{\linewidth}{!}{
    \Huge
    \begin{tabular}{lcccccc}
    \toprule[2.2pt]
Methods&  $\text{AP}_{Novel}$ & $\text{AP}_{Base}$ & $\text{AP}_{Mean}$ & $\text{AR}_{Novel}$ & $\text{AR}_{Base}$ & $\text{AR}_{Mean}$ \\
    \midrule[1.5pt]
DCMA & 7.53& 41.06& 14.82& 37.60& 69.39& 44.51 \\
CLIP-DCMA& 6.47& 40.27& 13.82& 45.37& 74.25& 51.65 \\
Box-DCMA &\textbf{9.17}& \textbf{42.04}& \textbf{16.31}& 43.16& 71.64& 49.35 \\
    \bottomrule[2.2pt]
    \end{tabular}
    }
    \vspace{3pt}
\end{table}

\subsubsection{Analysis of 3D Novel Object Discovery with Enrichment~(3D-NODE)}
\label{exp:3d-node}
\vspace{2mm}
\paragraph{Effect of 3D-NODE in CoDAv2}.
To further analyze the effect of our 3D-NODE, we integrate our Enrichment scheme into the CoDAv2 framework, as denoted by `3D-NODE + DCMA' in~\tabref{tab:main-ablation}. The comparison results clearly demonstrate substantial improvements with the integration of the Enrichment strategy over the CoDA configuration (`3D-NOD + DCMA'). In particular, the Enrichment strategy significantly enhances the $\text{AP}_{\text{Novel}}$~(7.53 vs. 6.71), proving its effectiveness in boosting model performance.
Additionally, the Enrichment strategy markedly increases the $\text{AR}_{\text{Novel}}$~(37.60 vs. 33.66), which confirms that the Enrichment strategy effectively extends the model's capability to detect more novel objects. This benefit is attributable to the inclusion of more varied and novel objects during the training phase, facilitated by the Enrichment process.

\begin{table}[!t]
    \centering
    \caption{Quantitative analysis of different IoU thresholds in the module Box-DCMA.}
    \label{tab:box_DCMA}    
    \newcommand{\tf}[1]{\textbf{#1}}
    \resizebox{\linewidth}{!}{
    \Huge
    \begin{tabular}{lcccccc}
    \toprule[2.2pt]
Methods &  $\text{AP}_{Novel}$ & $\text{AP}_{Base}$ & $\text{AP}_{Mean}$ & $\text{AR}_{Novel}$ & $\text{AR}_{Base}$ & $\text{AR}_{Mean}$ \\
    \midrule[1.5pt]
w/o Box-DCMA & 7.53& 41.06& 14.82& 37.60& 69.39& 44.51 \\
\updateblue{w/ Box-DCMA~(1e-3)}& \updateblue{8.40}& \updateblue{41.73}& \updateblue{15.65}& \updateblue{43.15}& \updateblue{71.35}& \updateblue{49.28} \\
w/ Box-DCMA~(5e-3)& \textbf{9.17}& \textbf{42.04}& \textbf{16.31}& 43.16& 71.64& 49.35 \\
w/ Box-DCMA~(5e-2)& 8.28& 41.15& 15.43& 41.04& 70.59& 47.47 \\
w/ Box-DCMA~(1e-2)& 8.71& 41.87& 15.92& 41.65& 71.37& 48.11 \\
    \bottomrule[2.2pt]
    \end{tabular}
    }
    \vspace{3pt}
\end{table}

\begin{table}[!t]
    \centering
    \caption{Quantitative analysis of inserting different numbers of novel objects into training 3D scenes by the proposed 3D Novel Object Enrichment strategy.}

    \label{tab:g_3dnod}    
    \newcommand{\tf}[1]{\textbf{#1}}
    \resizebox{\linewidth}{!}{
    \Large
    \begin{tabular}{lcccccc}
    \toprule[1.3pt]
Number&  $\text{AP}_{Novel}$ & $\text{AP}_{Base}$ & $\text{AP}_{Mean}$ & $\text{AR}_{Novel}$ & $\text{AR}_{Base}$ & $\text{AR}_{Mean}$ \\
    \midrule[1pt]
0& 6.71& 38.72& 13.66& 33.66& 66.42& 40.78 \\
3 & 7.19& 40.64& 14.46& 37.14& 67.99& 43.84 \\
5& \textbf{7.53}& \textbf{41.06}& \textbf{14.82}& 37.60& 69.39& 44.51 \\
7& 7.20& 40.62& 14.46& 37.38& 68.52& 44.15 \\
10& 7.29& 40.76& 14.56& 36.41& 68.82& 43.46 \\
    \bottomrule[1.3pt]
    \end{tabular}
    }
    \vspace{3pt}
\end{table}

\begin{figure}[!t]
     \centering
    \begin{overpic}[width=0.93\columnwidth]{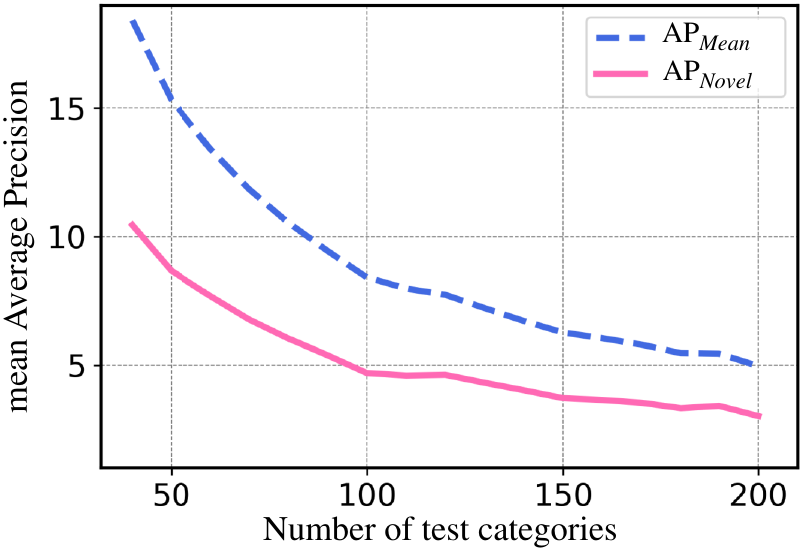}
    \end{overpic}
    \caption{AP results on SUN-RGBD~\cite{song2015sun} with the higher number of test categories (40 to 200). `$\text{AP}_{Mean}$' denotes mAP on all categories (i.e., base and novel categories), and `$\text{AP}_{Novel}$' denotes mAP on novel categories. The open-vocabulary ability is limited when a substantial number of novel categories are considered during testing. The main reason may be that pure point clouds are less discriminative than 2D color images. This presents a common challenge in 3D open-vocabulary scene understanding, as discussed in related works~\cite{peng2022openscene, ding2022language, liu2021language}.}
    \label{fig:vob_length}
\end{figure}

\paragraph{Number of Novel Objects Inserted into the Training Scenes in the Enrichment Strategy}. 
 We perform an ablation study on the Enrichment strategy in the extended method 3D-NODE, as presented in~\tabref{tab:g_3dnod}. This strategy involves introducing varying numbers of novel objects into the training 3D scenes, aiming to enrich the data distribution to varying degrees.
As illustrated in~\tabref{tab:g_3dnod}, when compared to the baseline where the Enrichment strategy is not used ( `0' in the first row of~\tabref{tab:g_3dnod}), all configurations with the Enrichment strategy demonstrate improvements. Notably, the configuration with five novel objects (the third row of~\tabref{tab:g_3dnod}) delivers the most significant improvements. This optimal number avoids the excessive introduction of noise while ensuring a substantial enrichment of novel objects. Consequently, we have adopted this setting (five novel objects) in our final methodology.

\begin{figure}[!t]
\centering
    \begin{overpic}[width=1\columnwidth]{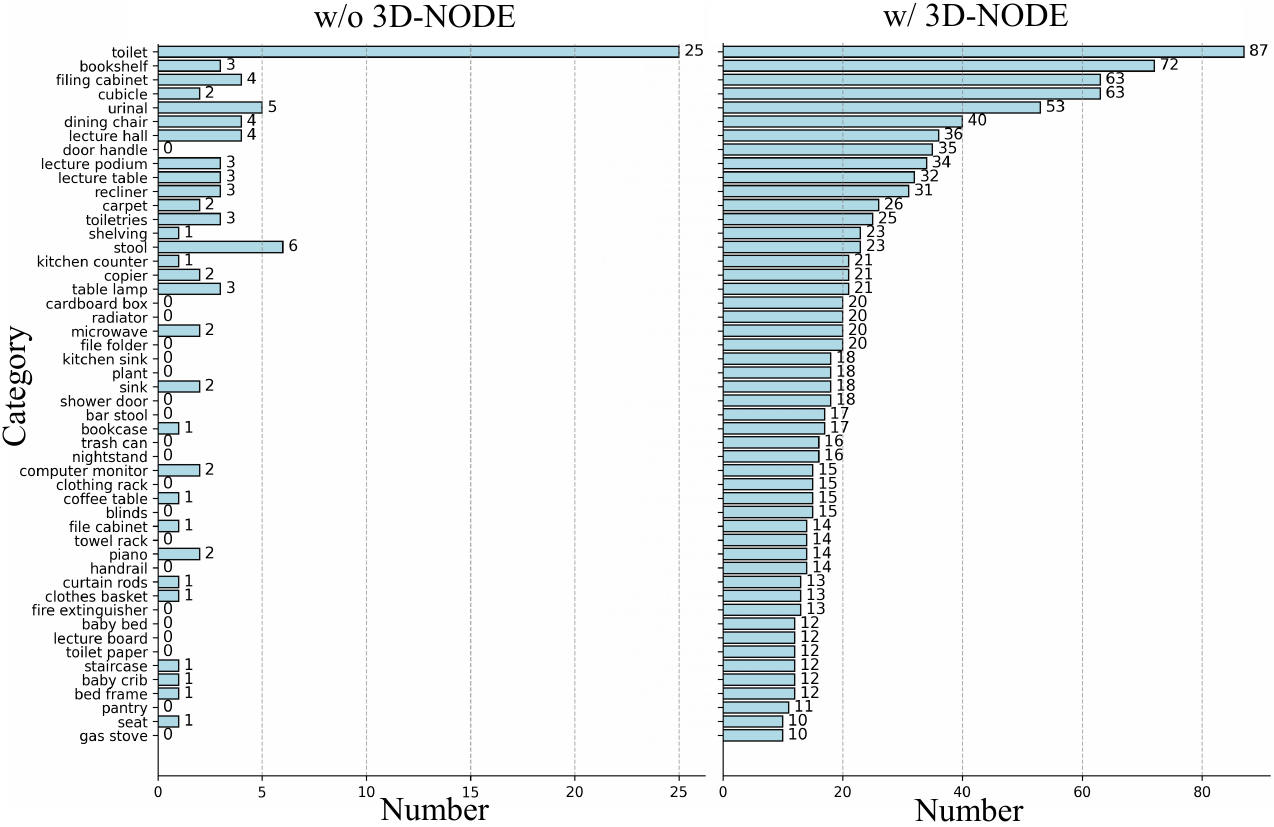}
    \end{overpic}
    \caption{\updateblue{Category distribution of the discovered novel objects in the training set of SUN-RGBD. As shown, training with our proposed 3D-NODE enables the model to discover more novel objects, which serve as training samples and provide supervision signals for novel categories, thereby alleviating the long-tail problem.}}
    \label{fig:cate_dis}
\end{figure}

\updateblue{
\paragraph{Category Distribution of the Discovered Novel Objects.}
 We counted the number of novel objects discovered by the model in the training set. 
These discovered objects serve as training samples and provide supervision signals for novel categories.
As shown in~\figref{fig:cate_dis}, training with our 3D-NODE enables the model to discover significantly more novel objects across various categories, effectively alleviating the long-tail problem. Specifically, as for the bottom 30\% categories in terms of frequency, the proportion of objects in these tail categories increased from 8.79\% to 15.57\%, representing a relative improvement of \textbf{77.13\%}. }

\subsubsection{Analysis of Discovery-driven Cross-modal Alignment with Box Guidance~(Box-DCMA)}
\label{exp:box-dcma}
\vspace{2mm}

\begin{table*}[!t]
    \centering
    \caption{Comparison with other alternative methods on the SUN-RGBD and ScanNetv2 datasets. The `inputs' in the second column indicate what inputs are used during the testing phase. During testing, For `3D-CLIP', 2D images are necessary to carry out open-vocabulary classification on detected objects. In contrast, our approach \textbf{CoDAv2} adopts only pure point clouds as inputs during testing. The proposed CoDAv2 establishes significant state-of-the-art results.}
    \label{tab:sota-sunrgbd}    
    \newcommand{\tf}[1]{\textbf{#1}}
    \resizebox{0.9\textwidth}{!}{
    \begin{tabular}{lccccccc}
    
    \toprule

Methods & Inputs &$\text{AP}_{Novel}$ & $\text{AP}_{Base}$ & $\text{AP}_{Mean}$ & $\text{AR}_{Novel}$ & $\text{AR}_{Base}$ & $\text{AR}_{Mean}$ \\
    \midrule
      \multicolumn{8}{c}{SUN-RGBD} \\
     \midrule
Det-PointCLIP~\cite{zhang2022pointclip} &Point Cloud & 0.09& 5.04& 1.17& 21.98& 65.03& 31.33 \\
Det-PointCLIPv2~\cite{zhu2022pointclip} &Point Cloud & 0.12& 4.82& 1.14& 21.33& 63.74& 30.55 \\
Det-CLIP$^2$~\cite{zeng2023clip2}         &Point Cloud & 0.88& 22.74& 5.63& 22.21& 65.04& 31.52 \\
3D-CLIP~\cite{radford2021learning} &Image\&Point Cloud & 3.61& 30.56& 9.47& 21.47& 63.74& 30.66 \\

CoDA~\cite{cao2023coda} &Point Cloud & 6.71 & 38.72 & 13.66 & 33.66 & 66.42 & 40.78 \\ 
\textbf{CoDAv2} &Point Cloud & \textbf{9.17} & \textbf{42.04} & \textbf{16.31} & \textbf{43.16} & \textbf{71.64} & \textbf{49.35} \\ 
    \midrule
    \multicolumn{8}{c}{ScanNetv2} \\
    \midrule
Det-PointCLIP~\cite{zhang2022pointclip} &Point Cloud &0.13 &2.38
&0.50 &33.38	&54.88 &36.96 \\
Det-PointCLIPv2~\cite{zhu2022pointclip} &Point Cloud &0.13 &1.75 &0.40 &32.60 &54.52 &36.25 \\
Det-CLIP$^2$~\cite{zeng2023clip2}         &Point Cloud & 0.14& 1.76& 0.40& 34.26& 56.22& 37.92 \\
3D-CLIP~\cite{radford2021learning} &Image\&Point Cloud &3.74 &14.14 &5.47 &32.15 &54.15 &35.81 \\
CoDA~\cite{cao2023coda} &Point Cloud & 6.54 &21.57 &9.04 &43.36 &61.00 &46.30 \\ 
\textbf{CoDAv2} &Point Cloud & \textbf{9.12} & \textbf{23.35} & \textbf{11.49} & \textbf{64.00} & \textbf{72.16} & \textbf{65.36} \\ 

    \bottomrule
    \end{tabular}
    }

\end{table*}

\begin{table*}[!t]
    \centering
    \caption{Quantitative comparison of methods under the same setting of~\cite{lu2023open} on ScanNetv2. The `Mean' column denotes the average performance across all 20 novel categories.}
    \label{tab:cmpwith3ddet}    
    \newcommand{\tf}[1]{\textbf{#1}}
    \resizebox{1\textwidth}{!}{
    \begin{tabular}{l|c|cccccccccc}
    \toprule
Methods& \textbf{Mean} & toilet              & bed                 & chair               & sofa                & dresser            & table               & cabinet            & bookshelf          & pillow              & sink               \\
\midrule
OV-3DET~\cite{lu2023open} & 18.02 & 57.29                        & 42.26                        & 27.06                        & 31.50                         & 8.21                        & 14.17                        & 2.98                        & 5.56                        & 23.00                           & 31.60                                                                      \\
  CoDA~\cite{cao2023coda}  & 19.32 & 68.09 & 44.04 &  28.72 &  44.57 &  3.41 & 20.23 &  5.32 &  0.03 &  27.95 &  45.26     \\
    \textbf{CoDAv2}  & \textbf{22.72} & 77.24 & 43.96 &  15.05 &  53.27 &  11.37 & 19.36 &  1.42 &  0.11 & 34.42 &  44.38     \\
  \midrule
 Methods & - & bathtub             & refrigerator       & desk                & nightstand          & counter            & door               &  curtain & box                & lamp               & bag                               \\
 \midrule
 OV-3DET~\cite{lu2023open} & - & 56.28                        & 10.99                       & 19.72                        & 0.77                         & 0.31                        & 9.59                        & 10.53                                   & 3.78                        & 2.11                        & 2.71    \\
 CoDA~\cite{cao2023coda}  & - & 50.51 &  6.55 &  12.42 &  15.15 &  0.68 &  7.95 &  0.01  &  2.94 &  0.51 &  2.02  \\
  \textbf{CoDAv2} & - &  55.60 &  24.41 &  20.67 &   20.72 &  0.28 &  13.54 & 0.92  &  4.16 &  4.37 &  9.20  \\
 
    \bottomrule
    \end{tabular}
        }
\end{table*}

\paragraph{Effect of Box-DCMA in CoDAv2}.
To study the impact of Box-DCMA, we incorporate our proposed box guidance technique into the DCMA framework, denoted as `3D-NODE + Box-DCMA' in Table~\ref{tab:main-ablation}. When compared with the baseline model trained without box guidance (`3D-NODE + DCMA'), the introduction of box guidance substantially enhances the performance. Specifically, we observe significant improvements in both $\text{AP}_{\text{Novel}}$~(9.17 vs. 7.53) and $\text{AR}_{\text{Novel}}$~(43.16 vs. 37.60), thereby confirming the effectiveness of the Box-DCMA strategy. Besides, We also provide
qualitative analysis in~\figref{fig:box_dcma}.
As we can observe, `Box-DCMA' reduces the detection of background boxes compared to `DCAM', attributed to the proposed box guidance.

\paragraph{Effect of Different Matching Strategies for the `Background' Category in Box-DCMA.}
We further conduct an ablation study on different matching methods in the `2D-Box-Guide BG Matching' of Box-DCMA as shown in \tabref{tab:assign_DCMA}. The entry `DCMA' in the first row of \tabref{tab:assign_DCMA} indicates the use of DCMA instead of Box-DCMA during training, i.e., without matching background boxes. 
`CLIP-DCMA' signifies the utilization of CLIP~\cite{radford2021learning} to match `background' boxes. Specifically, we incorporate `background' into the supercategory list $T^{super}$, as described in~\secref{sec:novel-object-discovery-with-joint-priors}. The semantic probability distribution is then estimated according to~\equref{equ:get_3d_prob}. If the category with the maximum probability is `background', we assign `background' text to the corresponding predictions in the subsequent class-specific contrastive alignment.
As observed, the $\text{AP}$ with `CLIP-DCMA' is lower than that achieved by the original DCMA, yet it achieves a higher $\text{AR}$. This outcome suggests that CLIP may often misclassify background regions as foreground categories, leading to an overclassification of boxes as foreground, which in turn results in higher $\text{AR}$ but lower $\text{AP}$. After integrating box guidance into DCMA, our enhanced Box-DCMA significantly boosts $\text{AP}$, demonstrating that the proposed box guidance effectively discriminates background areas.

\paragraph{IoU Threshold in Box-DCMA.}
We perform an ablation study to evaluate the impact of the IoU threshold in the Box-DCMA, as outlined in~\secref{sec:box-dcma}. The results, presented in~\tabref{tab:box_DCMA}, show that all configurations of the IoU threshold yield improvements over the baseline DCMA approach, which is detailed in the first row of~\tabref{tab:box_DCMA}. The optimal performance is achieved with a threshold setting of 5e-3.

\begin{figure*}[!t]
    \begin{overpic}[width=1\textwidth]{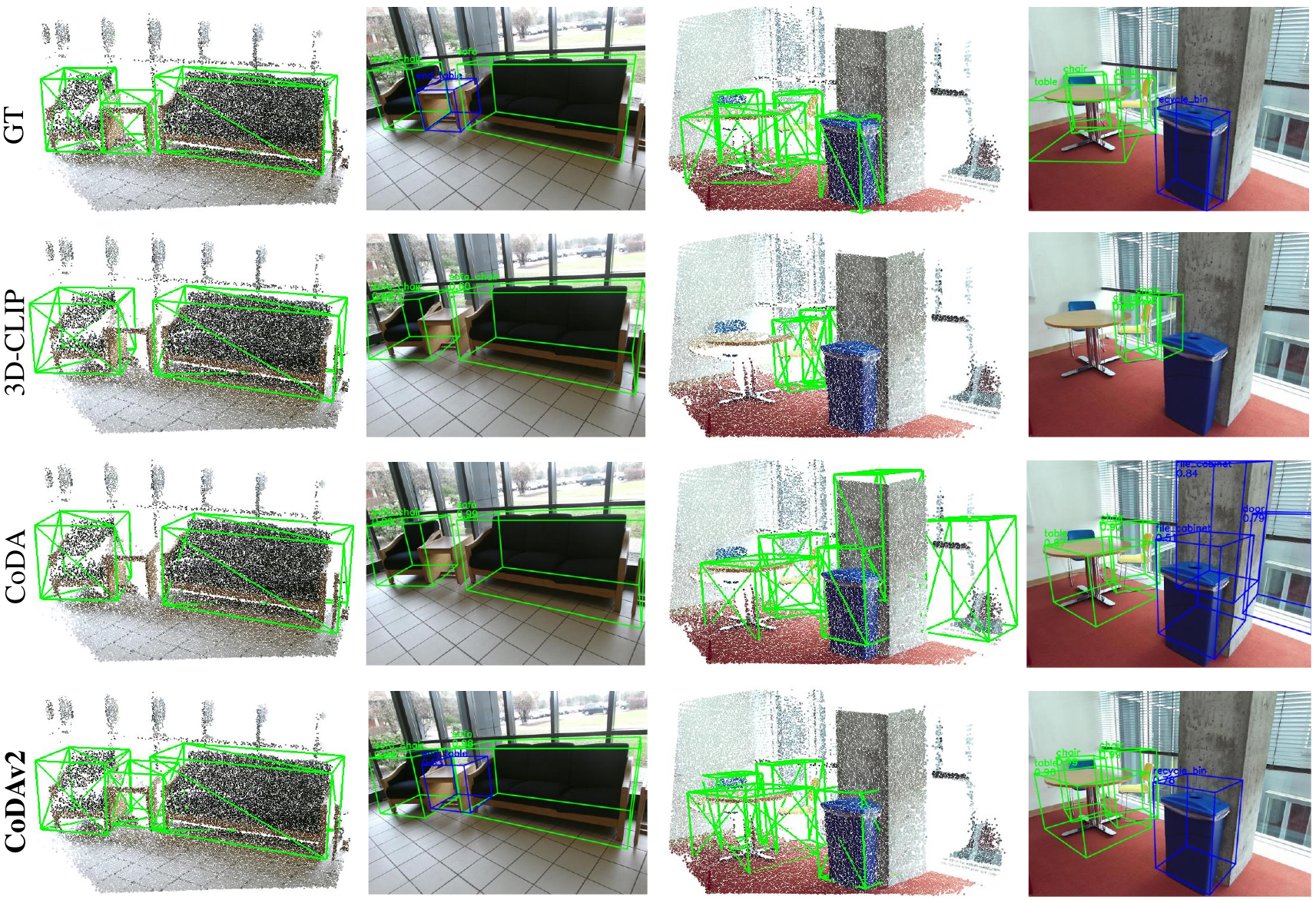}
    \end{overpic}
    \caption{Qualitative comparison with 3D-CLIP~\cite{radford2021learning} and CoDA~\cite{cao2023coda}. The first row shows the Ground Truth. Compared with the first two columns, it is evident that, enhanced by our 3D-NODE, our \textbf{CoDAv2} detects more novel objects, such as the end table highlighted by the blue boxes in the 2D color images. Comparing between the last two columns, benefiting from our Box-DCMA, \textbf{CoDAv2} manages to identify more novel objects~(e.g., the recycle bin labeled by the blue boxes in the 2D color images) while simultaneously avoiding the introduction of additional background noise boxes. (Zoom in for best view.)}
    \label{fig:cmp}
\end{figure*}

\subsubsection{Number of Test Object Categories in Evaluation} 
\label{sec:analysis_method}
\vspace{2mm}
In the ablation studies above, we evaluated our method on SUN-RGBD~\cite{song2015sun} with 46 classes, including 36 novel classes. To further assess the open-vocabulary capabilities of our approach, we increased the number of test categories from 40 to 200. As shown in~\figref{fig:vob_length}, as the number of test categories increases, $\text{AP}_{Mean}$ and $\text{AP}_{Novel}$ decrease to 4.96\% and 3.01\%, respectively.
Given that the inputs are pure point clouds, which are less discriminative than 2D color images, the open-vocabulary ability remains constrained as more categories are introduced. This is a common challenge in 3D open-vocabulary scene understanding~\cite{peng2022openscene, ding2022language, liu2021language}. Future work may incorporate multi-modality inputs to further enhance the performance on larger vocabulary.

\updateblue{\subsubsection{Discussion on Potential and Challenges}}~\label{sec:potential}
\updateblue{
\paragraph{Performance of Novel Categories.}
The present performance of novel categories is relatively limited, which is primarily caused by more false negatives (FN) compared to false positives (FP). Specifically, we analyzed the counts of FP and FN and observed that FN is 39\% higher than FP.
The main reason for higher FN is the insufficient number of training samples for novel objects, which limits the model's ability to detect these categories.
We also provide detailed statistics regarding the discovered novel objects in the pool. As shown in~\figref{fig:cate_dis}, while our model trained with 3D-NODE can effectively discover more novel objects than the model trained without 3D-NODE, the overall number of samples for tail categories remains limited. This scarcity of training samples in tail categories contributes to higher FN for these categories.
It is worth noting that this challenge is not unique to our method but is a common issue in the field, inherently stemming from the limitations of currently available datasets.
Nevertheless, our proposed method, which discovers more novel objects during training, has effectively alleviated this challenge, demonstrating significant advantages over the state-of-the-art method CoDA~\cite{cao2023coda} ($\text{AP}_{Novel}$: 9.17 vs. 6.71).}

\updateblue{
\paragraph{Utilization of 2D Box Guidance.}
Open-vocabulary 2D object detection~(OV-2DDet) method~\cite{liu2024grounding} might not perform well in some specific environments. However, in most scenes, OV-2DDet~\cite{liu2024grounding} demonstrates strong open-vocabulary capabilities. One important reason for this is the availability of much larger-scale training datasets for 2D detection (e.g., Objects365~\cite{shao2019objects365} with about 600k images for training). Such large-scale datasets enable OV-2DDet to learn richer open-vocabulary knowledge, whereas 3D datasets are significantly smaller in scale (e.g., SUN RGB-D with about 5k images for training).
Therefore, for OV-3DDet, OV-2DDet provides a valuable prior that complements its open-vocabulary capabilities. Motivated by this observation, we introduced the box guidance and designed Box-DCMA. Our experimental results validate both the motivation and the superiority of our method. As shown in~\tabref{tab:assign_DCMA}, the proposed Box-DCMA significantly outperforms DCMA in novel object detection, achieving higher $\text{AP}_{Novel}$ (9.17 vs. 7.53) and $\text{AR}_{Novel}$ (43.16 vs. 37.60). 
In summary, while OV-2DDet may have limitations in specific scenes, leveraging its strengths in most cases allows Box-DCMA to effectively improve OV-3DDet.
}

\updateblue{
\paragraph{Stronger Pre-trained Model than CLIP.} We replaced CLIP~\cite{radford2021learning} with the DFN~\cite{fangdata} pre-trained model in our method, resulting in improved performance on novel categories: $\text{AP}_{Novel}$ increased from 9.17 to 9.54, and $\text{AR}_{Novel}$ improved from 43.16 to 43.84. These results prove that our method, equipped with stronger pre-trained models, can better handle novel object categories.}

\updateblue{
\paragraph{Performance under Other Metrics.} We compare the F1-scores of our CoDAv2 with the state-of-the-art method, i.e., our NeurIPS conference version CoDA~\cite{cao2023coda}. As shown in~\tabref{tab: f1_cmp}, CoDAv2 consistently outperforms CoDA across all metrics, achieving notable gains of \textbf{2.35} in $\text{F}_{Novel}$, \textbf{4.19} in $\text{F}_{Base}$, and \textbf{2.89} in $\text{F}_{Mean}$.}

\begin{table}[!h]
    \centering
    \caption{\updateblue{Comparison of F1-scores with CoDA~\cite{cao2023coda}~(state-of-the-art) on the SUN-RGBD dataset. 
    CoDAv2 achieves consistent improvements over CoDA across all metrics, with significant gains of +2.35 in $\text{F}_{Novel}$, +4.19 in $\text{F}_{Base}$, and +2.89 in $\text{F}_{Mean}$.
    Note that F1-scores are scaled the same as mAP (multiplied by 100) for consistency in representation.}}
    \label{tab: f1_cmp}    
    \newcommand{\tf}[1]{\textbf{#1}}
    \resizebox{0.95\linewidth}{!}{
    \begin{tabular}{lccc}
    \toprule[0.8pt]
\updateblue{Methods} &  \updateblue{$\text{F}_{Novel} $} & \updateblue{$\text{F}_{Base} $} & \updateblue{$\text{F}_{Mean} $} \\
    \midrule[0.6pt]
\updateblue{CoDA~\cite{cao2023coda}} & \updateblue{7.53}& \updateblue{41.57}& \updateblue{14.79} \\
\updateblue{CoDAv2} & \updateblue{\textbf{9.88~\small{{(+2.35)}}}}& \updateblue{\textbf{45.76~\small{{(+4.19)}}}} & \updateblue{\textbf{17.68~\small{{(+2.89)}}}} \\
    \bottomrule[0.8pt]
    \end{tabular}
    }
    \vspace{3pt}
\end{table}

\subsection{Comparison with Alternatives} 
\textbf{Quantitative Comparison.}
Open-vocabulary 3D object detection is still a very new problem. Given the scarcity of related works in the literature and the novelty of our open-vocabulary setting, there are no existing results that can be directly compared in our open-vocabulary setting. Thus, we adapt open-vocabulary point cloud classification methods to our setting and evaluate their effectiveness. Specifically, we employ 3DETR to generate pseudo boxes and utilize PointCLIP~\cite{zhang2022pointclip}, PointCLIPv2~\cite{zhu2022pointclip}, and Det-CLIP$^2$~\cite{zeng2023clip2} to perform open-vocabulary object detection. These implementations are referred to as `Det-PointCLIP', `Det-PointCLIPv2', and `Det-CLIP$^2$', respectively, in~\tabref{tab:sota-sunrgbd}.
Furthermore, we introduce a pipeline that uses 2D color images as input. This pipeline employs 3DETR to create pseudo 3D boxes, which are then projected onto 2D planes using the camera matrix to create 2D boxes. These boxes are subsequently used to crop the corresponding 2D regions for classification by the CLIP model~\cite{radford2021learning}. This method, denoted as `3D-CLIP' in~\tabref{tab:sota-sunrgbd}, allows us to generate 3D detection results. \tabref{tab:sota-sunrgbd} demonstrates that, even when only the 3D point clouds are utilized as inputs during the testing stage, our methods, CoDA and CoDAv2, can achieve significantly higher $\text{AP}_{\text{Novel}}$ and $\text{AR}_{\text{Novel}}$ than other comparison methods, illustrating the superiority of our designs in both novel object localization and classification.
In the SUN-RGBD dataset, CoDAv2 notably surpasses the best-performing alternative method (3D-CLIP) by more than \textbf{150\%}. Similarly, in the ScanNetv2 dataset, CoDAv2 outperforms the leading alternative (3D-CLIP) by more than \textbf{140\%}. These substantial margins fully demonstrate the efficacy of our approach.

\paragraph{Comparison with OV-3DET~\cite{lu2023open}}.
Notably, the setting in our method significantly differs from that of OV-3DET~\cite{lu2023open}. In our approach, the model is trained with annotations from a limited set of base categories (10 categories) and learns to discover novel categories during the training process. We then evaluate our model across a broader array of categories, specifically 60 in ScanNetv2 and 46 in SUN-RGBD. In contrast, the approach~\cite{lu2023open} leverages a large-scale pretrained 2D open vocabulary detection model, namely the OV-2DDet model~\cite{zhou2022detecting}, to generate pseudo labels for novel categories and conduct evaluations on only 20 categories. Due to these differences in settings and the foundational models used, drawing a direct comparison with~\cite{lu2023open} is not straightforward.
To facilitate a fair comparison, thanks to the released codes from~\cite{lu2023open} to generate pseudo labels on ScanNetv2, we can retrain our method under the same setting as~\cite{lu2023open} on ScanNetv2. The comparison is presented in~\tabref{tab:cmpwith3ddet}. As shown, our CoDAv2 model achieves a significantly improved mean AP, with an enhancement of~\textbf{4.7 points} across all categories, thereby validating the superiority of our approach.

\paragraph{Qualitative Comparison.}
As depicted in~\figref{fig:cmp}, upon the comparison between samples in the first two columns, it can be observed that, with the enhancement provided by our 3D-NODE, \textbf{CoDAv2} is capable of detecting more novel objects, e.g., the end table highlighted by the blue boxes in the 2D color images. In the analysis of the last two columns, thanks to our Box-DCMA, \textbf{CoDAv2} not only identifies more novel objects, such as the recycle bin marked by the blue boxes, but also effectively avoids the introduction of additional background noise boxes.

\section{Conclusion}\label{sec:conclusion}
In this paper, we propose a unified framework, coined as CoDAv2, to tackle the fundamental challenges in OV-3DDet, which targets simultaneous localization and classification for novel objects. To achieve localization for 3D novel objects, we introduce the 3D Novel Object Discovery~(3D-NOD) strategy, which leverages 3D geometry priors and 2D open-vocabulary semantic priors to enhance the discovery of novel objects during training. For the classification of novel objects, we develop a Discovery-driven Cross-Modal Alignment module~(DCMA). This module incorporates both class-agnostic and class-discriminative alignments to synchronize features across 3D, 2D, and text modalities.
Step forward, we enhance our contributions by introducing 3D Novel Object Enrichment and incorporating box guidance. The 3D Novel Object Enrichment strategy actively incorporates newly discovered novel objects into the training 3D scenes, thereby enhancing the model's capability to detect novel objects, while the box guidance is designed within the DCMA to improve the differentiation between background and foreground objects.
Benefiting from our contributions,
our CoDAv2 significantly outperforms the best-performing alternative method by more than \textbf{140\%} on two challenging datasets, i.e., SUN-RGBD and ScanNetv2.

\ifCLASSOPTIONcaptionsoff
  \newpage
\fi

\bibliographystyle{ieeetr}
\bibliography{tpami}

\end{document}